\documentclass{article}
\PassOptionsToPackage{numbers, sort&compress, comma, square}{natbib}
\usepackage[final]{neurips_2024}

\usepackage{color}
\usepackage{bm}
\usepackage{hyperref}
\usepackage{amsmath,amsfonts}
\usepackage{blindtext}
\usepackage{listings}
\usepackage{mathtools}

\DeclarePairedDelimiterX{\infdivx}[2]{(}{)}{%
	#1\;\delimsize\|\;#2%
}

\newcommand{\vect}[1]{\boldsymbol{\mathbf{#1}}}

\newcommand{\x}{\xv}
\newcommand{\y}{\yv}
\newcommand{\z}{\zv}

\newcommand{\dm}{\mathrm{d}}

\newcommand{\N}{\mathcal{N}}

\newcommand{\dt}{\mathrm{d}t}

\newcommand{\epsilonv}{\vect\epsilon}

\newcommand{\xv}{\vect x}
\newcommand{\yv}{\vect y}
\newcommand{\zv}{\vect z}

\newcommand{\Iv}{\vect I}

\newcommand{\Oc}{\mathcal O}

\usepackage[utf8]{inputenc} 
\usepackage[T1]{fontenc}    
\usepackage{hyperref}       
\usepackage{url}            
\usepackage{booktabs}       
\usepackage{amsfonts}       
\usepackage{nicefrac}       
\usepackage{microtype}      
\usepackage{xcolor}         
\usepackage{capt-of}
\usepackage{amsmath}
\usepackage{amssymb}
\usepackage{bbm}
\usepackage{mathtools}
\usepackage{amsthm}
\usepackage{physics}
\usepackage{multirow}
\usepackage{graphicx}
\usepackage{subfigure}
\usepackage{xspace}
\usepackage{tcolorbox}
\usepackage{wrapfig}
\usepackage[capitalize,noabbrev]{cleveref}
\tcbuselibrary{breakable}
\theoremstyle{plain}

\theoremstyle{definition}

\theoremstyle{remark}

\newcommand{\ourmodelmpdot}{DoT$^{\text{MP}}$\xspace}
\newcommand{\ourmodeldot}{DoT\xspace}

\usepackage[textsize=tiny]{todonotes}
\title{Diffusion of Thought: Chain-of-Thought Reasoning \\in Diffusion Language Models}

\author{%
Jiacheng Ye$^{1}$\footnotemark[1], 
  Shansan Gong$^{1}$\footnotemark[1],
  Liheng Chen$^{1}$\thanks{Equal contribution.}, 
  Lin Zheng$^{1}$, \\ \bf Jiahui Gao$^2$,Han Shi$^2$,Chuan Wu$^{1}$,Xin Jiang$^2$, Zhenguo Li$^2$,Wei Bi$^3$,Lingpeng Kong$^{1}$ \\
  $^1$ The University of Hong Kong \quad $^2$ Huawei Noah's Ark Lab \quad $^3$ Tencent AI Lab \\ 
  \texttt{\small \{carsonye, sansa933\}@connect.hku.hk}
}

\begin{document}

\maketitle

\begin{abstract}
Recently, diffusion models have garnered significant interest in the field of text processing due to their many potential advantages compared to conventional autoregressive models.
In this work, we propose Diffusion-of-Thought (DoT),  a novel approach that integrates diffusion models with Chain-of-Thought, a well-established technique for improving the reasoning ability of autoregressive language models. In contrast to autoregressive language models that make decisions in a left-to-right, token-by-token manner, DoT allows reasoning steps to diffuse over time through a diffusion language model and offers greater flexibility in trading-off computation for reasoning performance. Our experimental results demonstrate the effectiveness of DoT in multi-digit multiplication, boolean logic, and grade school math problems, with a small diffusion model outperforming a much larger autoregressive model in both efficiency and accuracy. In addition to that, DoT showcases promising self-correction abilities and benefits from existing reasoning-enhancing techniques like self-consistency decoding. Our findings contribute to the understanding and development of reasoning with diffusion language models.
\end{abstract}

\section{Introduction}

Large language models (LLMs) have had a profound impact on the entire field of artificial intelligence~\citep{gpt4,touvron2023llama}, transforming our approach to addressing classical problems in natural language processing and machine learning. Among the most notable aspects of LLMs is their remarkable reasoning ability, which many researchers consider to be a representative emergent capability brought about by LLMs~\citep{wei2022emergent}.
Chain-of-thought prompting (CoT)~\citep{Wei2022ChainOT}), which generates a series of intermediate reasoning steps in autoregressive (AR) way, has emerged as a central technique to support complex reasoning processes in LLMs.
Despite advancements, errors in intermediate CoT steps can lead to inaccurate answers ~\citep{lanham2023measuring}, posing self-correction difficulties ~\citep{huang2023large}, and concerns about CoT's inefficiency have been highlighted in recent studies~\citep{deng2023implicit}.

Recently, diffusion models have attracted interest in text processing~\citep{li2022diffusion, Zheng2023ARD,zou2023survey} as a result of success in the vision domain and distinctive modeling strengths over autoregressive models~\cite{lin2020limitations}, offering potential benefits including global planning ability~\citep{zhang2023planner, ye2023diffusion}, self correction~\citep{hoogeboom2021argmax} and efficiency~\citep{lou2023discrete}. As part of the research community effort, pre-trained diffusion language models such as Plaid~\citep{gulrajani2023likelihood} and SEDD~\citep{lou2023discrete} have shown significant progress in text generation capabilities. 
Although they have not yet attained the scale and capabilities of existing proprietary autoregressive LLMs like GPT-4~\citep{gpt4}, these models have demonstrated performance on par with GPT2~\citep{Brown2020LanguageMA} and the scaling law~\citep{kaplan2020scaling} in diffusion language models have been highlighted in Plaid.
As a result, it becomes pertinent to explore the following question: 
\textit{can diffusion language models also leverage the CoT-style technique to gain enhanced complex reasoning abilities?}

\begin{wrapfigure}{r}{0.5\textwidth}
    \vspace{-5pt}
    \begin{center}
    \includegraphics[width=0.9\linewidth]{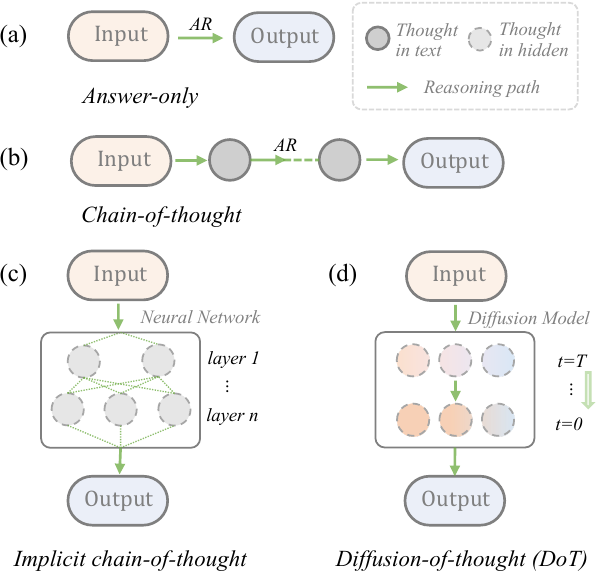}
    \end{center}
    \caption{Illustration of reasoning approaches. (a) \textbf{Answer-only} and (b) \textbf{CoT} generate left-to-right tokens by prompting autoregressive language model. (c) \textbf{Implicit CoT} replaces horizontal reasoning (CoT) with vertical reasoning from shallow layer to deep layer~\citep{deng2023implicit}. (d) \textbf{DoT} generates reasoning path along with the diffusion timesteps.}
    \label{fig:connect-dot}
    \vspace{-10pt}
\end{wrapfigure}

This work presents a preliminary study on this question. We propose Diffusion of Thought (DoT), an inherent chain-of-thought method tailored for diffusion models. In essence, DoT progressively updates a sequence of latent variables representing thoughts in the hidden space, allowing reasoning steps to diffuse over time in parallel. We also introduce a multi-pass variant of DoT which focuses on generating one thought at a time to compensate for causal bias.
To condition on complex queries, instead of using gradient-based classifier guidance~\citep{li2022diffusion,gulrajani2023likelihood}, DoT trains and samples from the denoising model using the classifier-free guidance as in~\citet{gong2022diffuseq}, to provide more reliable controlling signals on exact tokens.

Furthermore, to improve the self-correcting capability of the diffusion model, DoT integrates training-time sampling algorithms to learn to recover from errors originating from prior or current reasoning steps. This feature offers a fresh angle on the issue of error accumulation~\citep{lanham2023measuring,huang2023large} inherent in autoregressive models. 
Finally, we adapt a conditional ODE Solver~\citep{lu2022dpm++} for DoT during inference time to accelerate the inference of continuous diffusion models. We show DoT enjoys flexibility in trading off computation (reasoning time) and performance as more complex problems may necessitate increased computation in reasoning~\citep{banino2021pondernet,Wei2022ChainOT}.

From a methodological standpoint, DoT shares similarities with the recently proposed Implicit CoT approach~\citep{deng2023implicit}, where the latter learns thoughts in hidden states across transformer layers to improve the time efficiency of autoregressive CoT generation.
A schematic illustration of CoT, Implicit CoT, and DoT can be found in Figure~\ref{fig:connect-dot}. 

The main contributions of our paper are threefold:
\begin{enumerate}
    \item We first introduce the reasoning technique for diffusion models (DoT), and showcase its advantages in simple reasoning tasks (digit multiplication and boolean logic) when compared to autoregressive CoT and Implicit CoT. DoT achieves up to $27\times$ speed-up without performance drop (\S\ref{sec:exp-main-digit}).
    \item  We further adapt DoT to continuous and discrete diffusion base models, and introduce two training-time sampling algorithms to improve its self-correction ability. DoT exhibits superior performance compared to GPT2 with CoT on grade school math problems, enabling a small diffusion model to outperform a 4.6x larger autoregressive model, showing the potential of text diffusion models for complex reasoning (\S\ref{sec:exp-main-gsm}).
    \item Our analysis demonstrates the flexibility of DoT in the trade-off between reasoning time and performance (\S\ref{sec:exp-trade-off}), and showcases DoT's self-correction capability (\S\ref{sec:exp-self-corr}). We also find that self-consistency decoding can further improve DoT and its multi-pass variant (\S\ref{sec:exp-self-consis}).
\end{enumerate}
Although it is challenging for current pre-trained diffusion language models to directly compete with LLMs that are hundreds of times larger in parameter size, our study emphasizes the possibility of their complex reasoning abilities and highlights the substantial potential in developing LLMs that go beyond the autoregressive paradigm. 
We release all the codes at \href{https://github.com/HKUNLP/diffusion-of-thoughts}{https://github.com/HKUNLP/diffusion-of-thoughts}.

\section{Preliminaries}
This section introduces key concepts and notations in diffusion models for text generation. Detailed formulations and derivations are provided in Appendix~\ref{sec:app-eq}.

A typical diffusion model contains the forward and reverse process. For each forward step $ q(\mathbf{z}_t|\mathbf{z}_{t-1})$, we gradually inject noise into the data representation $\mathbf{z}_{t-1}$ 
 from the last timestep to obtain $\mathbf{z}_{t}$. Here $t = 1, 2,...,T$ and the larger $t$ corresponds to noisier data. For reverse process, the ultimate goal is to recover the original $\mathbf{z}_0$ by denoising $\mathbf{z}_t$: $p_{\theta}(\mathbf{z}_{0:T}):=p(\mathbf{z}_T)\prod_{t=1}^Tp_{\theta}(\mathbf{z}_{t-1}|\mathbf{z}_t)$. We model the learning process $p_{\theta}(\mathbf{z}_{t-1}|\mathbf{z}_t)$ using the proposed diffusion model $\z_{\theta}(\mathbf{z}_t, t)$.

Previous text generation using diffusion models almost contains two categories: (1) Continuous diffusion models such as Diffusion-LM~\citep{li2022diffusion}, which relies on a mapping function between the real values and feasible integral point; (2) Discrete diffusion models like D3PM~\citep{austin2021structured}, which directly formulate the problem as the integer program.
Continuous diffusion models map the discrete text $\mathbf{w}$ into a continuous space through an embedding function $\textsc{Emb}(\mathbf{w})$, and its inverse operation is called rounding. The forward perturbations are applied according to $q(\mathbf{z}_{t} \vert \mathbf{z}_{t-1}) = \mathcal{N}(\mathbf{z}_{t};\sqrt{1-\beta_t}\mathbf{z}_{t-1}, {\beta}_t \mathbf{I})$, where $\beta_t \in (0,1)$ represents different scales of the Gaussian noise. Plaid~\citep{gulrajani2023likelihood} is a continuous diffusion language model trained from scratch on 314B tokens with $1024$ context size. It is currently the largest scale diffusion language model with 1.3B parameters. In the case of discrete diffusion models, each $\mathbf{z}_t$ is represented as a discrete random variable using one-hot vectors in $\{0, 1\}^K$, where $K$ denotes the vocabulary size. They define $q(\mathbf{z}_{t} \vert \mathbf{z}_{t-1})$ through a transition matrix, making it a point mass with probability on an absorbing state \texttt{[MASK]} or a uniform distribution over the vocabulary size. SEDD~\citep{lou2023discrete} is a recently trained-from-scratch discrete diffusion language model with small and medium size similar to GPT2.

For sequence-to-sequence (seq2seq) generation, which involves a pair of sequences $\mathbf{w}^x$ and $\mathbf{w}^y$, DiffuSeq~\citep{gong2022diffuseq} treats these two sequences as a single one $\mathbf{w}^{z}=\mathbf{w}^{[x; y]}$ and uses a left-aligned mask $[\mathbf{0};\mathbf{1}]$ during the forward and reverse diffusion process to distinguish them. Unlike traditional diffusion models that corrupt the entire $\mathbf{z}_{t}$, DiffuSeq only adds noise to those entries with the mask value of 1 (e.g., $\mathbf{y}_t$). This modification, termed partial noising, tailors diffusion models for conditional language generation, and set a difference between the gradient-based token guidance in~\citep{li2022diffusion} and \citep{gulrajani2023likelihood}.

\section{Diffusion-of-Thoughts}

\begin{figure*}[!t]
    \centering
    \includegraphics[width=1\linewidth]{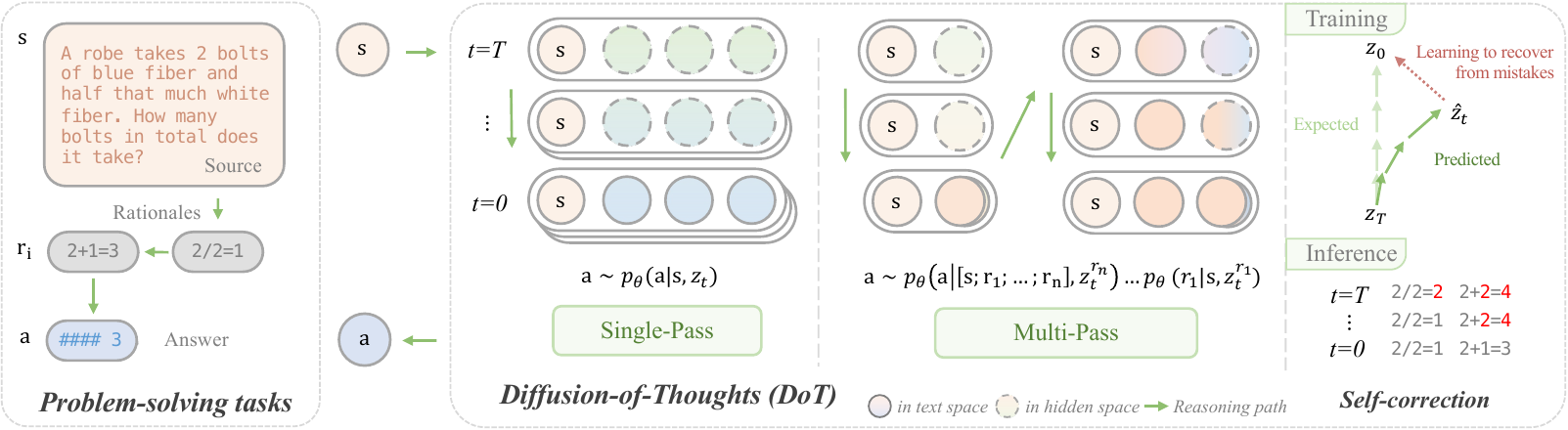}
    \caption{Demonstration of DoT pipeline. DoT diffuses all possible thoughts across diffusion timestep $t$. Multi-pass DoT disentangles each rationale and introduces causal bias. The stacked circles stand for the marginalization over other potential reasoning paths, which is implicitly carried out during the training of diffusion models. }
    \label{fig:pipe-dot}
    \vspace{-5pt}
\end{figure*}

In this section, 
we begin with an overview of our method and its relationship with other reasoning paradigms (\S\ref{sec:pro-state}). 
We then introduce Diffusion-of-Thoughts (DoT) as well as its multi-pass variant (\ourmodelmpdot; \S\ref{sec:dot}), as illustrated in Figure~\ref{fig:pipe-dot}. Following this, we outline the implementation of our training (\S\ref{sec:training}) and inference (\S\ref{sec:inference}) protocols.

\subsection{Overview}
\label{sec:pro-state}

Without loss of generality, we use the mathematical problem-solving task as our running example. A problem statement and its correct answer are denoted as $\mathbf{s}$ and $\mathbf{a}$, respectively. We employ a language model with parameters $\theta$, represented as $p_\theta^{\textit{LM}}$, to find the solution for each problem. For regular usage of language models without Chain-of-Thoughts (CoT), the final answer $\mathbf{a}$ is generated directly as $\mathbf{a}\sim p_\theta^{\textit{LM}}(\mathbf{a}|\mathbf{s})$. The CoT approach introduces meaningful intermediate steps or rationales $\mathbf{r}_1, \dots, \mathbf{r}_n$ for language models to bridge $\mathbf{s}$ and $\mathbf{a}$, resulting in the output $\mathbf{a}\sim p_\theta^{\textit{LM}}(\mathbf{a}|\mathbf{s}, \mathbf{r}_{1\dots n})$. 
For implicit CoT~\citep{deng2023implicit}, the hidden representations of rationales $\mathbf{z}_1, \dots, \mathbf{z}_n$ are distilled into transformer layers, leading to $\mathbf{a}\sim p_\theta^{\textit{iCoT}}(\mathbf{a}|\mathbf{s}, \mathbf{z}_{1\dots n})$. Similarly but differently, for DoT, these representations are distributed over diffusion timestep $t$ as $\mathbf{a}\sim p_\theta^{\textit{DoT}}(\mathbf{a}|\mathbf{s}, \mathbf{z}_t)$, where $\mathbf{z}_t$ corresponds exactly to the noised data in diffusion models.

\subsection{Modeling}
\label{sec:dot}
We begin by observing the gradient-based token guidance fails to do accurate conditioning as the model cannot exactly recover each conditioning token (see Table~\ref{tab:ablation-gsm}). This is vital, especially in mathematical reasoning, as it is expected to perform reasoning based on exact tokens (e.g., numbers) in the problem statement, rather than more compact gradient signals.
For this, we adopt DiffuSeq-style~\citep{gong2022diffuseq} classifier-free conditioning during the fine-tuning of Plaid, where all rationales are generated by the backward diffusion process in parallel, with all the conditional tokens fixed as still. Specifically, the problem context $s$ is concatenated with the rationales $\mathbf{r}_{1\dots n}$ during training and sampling, while the noise is only partially imposed to the rationale part in $\mathbf{r}_{1\dots n}$, keeping $\mathbf{s}$ anchored as the condition.

We further propose a multi-pass (MP) variant of DoT, denoted as \ourmodelmpdot, which generates rationales in a thought-by-thought paradigm. This method disentangles the generation of multiple rationales and introduces casual inductive bias such that later rationale can be guided by stronger condition signals of prior rationales during the generation. Specifically, in the first pass, we generate the first rationale by $\mathbf{r}_1\sim p_{\theta}^{\textit{DoT}}(\mathbf{r}_1|\mathbf{s}, \mathbf{z}^{r_1}_t)$, where $\mathbf{z}^{r_1}_t$ is the noised vector representation of $\mathbf{r}_1$ in diffusion model. Then $\mathbf{r}_1$ is connected to $\mathbf{s}$ as the condition $[\mathbf{s};\mathbf{r}_1]$ to get $\mathbf{r}_2\sim p_{\theta}^{\textit{DoT}}(\mathbf{r}_2|[\mathbf{s};\mathbf{r}_1], \mathbf{z}^{r_2}_t)$, and then we have $[\mathbf{s};\mathbf{r}_1;\mathbf{r}_2]$. Through multiple iterations, we can get the final answer: $\mathbf{a}\sim p_{\theta}^{\textit{DoT}}(\mathbf{a}|[\mathbf{s};\mathbf{r}_1;...;\mathbf{r}_n], \mathbf{z}^{r_n}_t)$.

\subsection{Training}
\label{sec:training}
\paragraph{Scheduled sampling}
Diffusion models have intrinsic self-correcting capability through the multi-step denoising process. To further improve their self-correcting ability, we design a \textit{scheduled sampling}~\citep{Bengio2015ScheduledSF} mechanism tailored for diffusion models such that self-generated error thoughts in previous timesteps are exposed and corrected during the training stage. 
Formally, for any timesteps $s$, $t$, $u$ that satisfy $1<s<t<u<T$, $\mathbf{z}_t$ is sampled from the forward distribution $q\left(\mathbf{z}_t \mid \mathbf{z}_0\right)$ in the training stage while during inference it is sampled from $q(\mathbf z_t \mid  \z_\theta \left(\mathbf{z}_u ; u\right))$ instead, where $\z_{{\theta}}$ is a denoiser neural network that reparameterizes $\mathbb{E}_q[\mathbf{z}_0|\mathbf{z}_t]$. The presence of such exposure bias may impede the model's ability to recover from erroneous thoughts during the generation process as the model $\z_{{\theta}}$ has only been trained on corruptions $\mathbf{z}_t$ diffused from oracle data.
To mitigate this problem, we mimic the inference stage with probability $\epsilon_i$ during training depending on the current training step $i$, and $\epsilon_i$ linearly decays from 1 to $\epsilon_{min}$. Specifically, for time-step $t$, we randomly sample a former time-step $u\in\{t+1,\dots, T\}$, obtain $\mathbf z_u$ by forward noising and perform a model forward pass to get a predicted $\hat{\mathbf z}_0=\z_{\theta}\left(\mathbf{z}_u ; u\right))$. $\mathbf z_t$ is then sampled from $q(\mathbf z_t \mid  \hat{\mathbf z}_0)$ to replace the regular one in loss calculation.
Compared with scheduled sampling for autoregressive models, such a mechanism in DoT helps the model to recover from errors by considering global information instead of relying on the left-side tokens.

\paragraph{Coupled sampling}

In \ourmodelmpdot, correct previous thoughts are given in the training stage, which is not given during inference. 
Similar to auto-regressive decoding, \ourmodelmpdot may suffer from error accumulation during the thought-by-thought generation process. To enhance the self-correction ability of \ourmodelmpdot, we propose a coupled sampling mechanism by adding noise not only to the current thought but also to previous thoughts during training with some probability.
For instance, the previous sequence $\mathbf{z}_0=\textsc{Emb}([\mathbf{s};\mathbf{r}_{1}])$ will be modified to $\mathbf{z}_0=\textsc{Emb}([\mathbf{s};\mathbf{r}_{1};\mathbf{r}_{2}])$, with the partial noise being applied to $[\mathbf{r}_{1};\mathbf{r}_{2}]$ rather than just the last rationale $\mathbf{r}_{2}$. 
Therefore, the model learns to be robust to errors in $\mathbf{r}_1$ when predicting $\mathbf{r}_2$, which better aligns with the inference stage. The new $\mathbf{z}_0$ will be reparameterized into $\mathbf{z}_t$ as before and other procedures keep the same.

\paragraph{Training objective}
Given a set of training data for problem-solving tasks of size $D$: $\{\mathbf{s}^j, \mathbf{r}^j_{1\dots n}, \mathbf{a}^j\}_{j\in D}$, we have two training settings for DoT models: one is training from scratch, while the other is fine-tuning from the pre-trained diffusion model. In both training settings, we share the same training objective. For example, the objective is to minimize the negative variational lower bound $\mathcal{L}_{\text{VLB}} (\mathbf{w}^z)$ in continuous diffusion models:
\begin{equation}
\begin{aligned}
\mathcal{L}_{\text{VLB}}(\mathbf{w}^z)=&\mathbb{E}_{q({\mathbf{z}_0}\mid \mathbf{w}^z)} \Bigg[ \underbrace{\log\frac{ q(\mathbf{z}_T|\mathbf{w}^z)}{p_{\theta}(\mathbf{z}_T)}}_{\text{Prior loss}}
+ \underbrace{\textstyle \mathcal{L}_{\text{VLB}}(\mathbf{z}_0)}_{\text{Diffusion loss}} 
 \underbrace{-\log p_\theta(\mathbf{w}^z|\mathbf{z}_0)}_{\text{Rounding loss}}\Bigg],
\end{aligned}
\end{equation}
where the rounding loss regularizes the embedding learning and the diffusion loss sums up the KL divergence of each time step $t$ with different weighting terms. Please refer to Appendix~\ref{sec:app-eq} for a detailed training objective formulation of continuous and discrete diffusion models. 

\subsection{Inference}
\label{sec:inference}
One of the significant advantages of diffusion models is their inference flexibility. Naturally, more complex problems may necessitate increased computation in reasoning time~\citep{banino2021pondernet,Wei2022ChainOT}, which can be controlled by setting a larger backward timestep $T$ in DoT. However, continuous diffusion such as Plaid usually requires more timesteps, e.g., 4096~\citep{gulrajani2023likelihood}, to converge. To accelerate the sampling process of the continuous diffusion, we adapt the ODE Solver~\citep{lu2022dpm++,lu2022dpm} into a conditional form to fit the conditional training process (detailed in Appendix~\ref{app:conditional-ode-solver}).
Moreover, sharing a similar idea of MBR~\citep{koehn2004statistical}, self-consistency~\citep{wang2022self} boosts the performance of CoT significantly by generating and aggregating multiple samples. 
In the context of diffusion models, we can also expect its potential improvement using self-consistency, thanks to their ability to naturally produce diverse responses~\citep{gong2022diffuseq}. 
After sampling $m$ times to obtain multiple reasoning pathways $(\mathbf{r}_{i;1...n}, \mathbf{a}_i)$ from DoT, self-consistency involves marginalizing over $\mathbf{r}_{i;1...n}$ by taking a majority vote over $\mathbf{a}_i$, i.e., 
$\arg \max_{\mathbf{a}}\sum_{i=1}^{m} \mathbbm{1}(\mathbf{a}_i = a)$. We consider this as the most ``consistent'' answer among the candidate set of $m$ answers. 

\section{Experiments}
We conduct experiments on both simple multi-digit multiplication and boolean logic reasoning as well as complex grade school math problems, to explore the reasoning paradigm in diffusion models.
\subsection{Experimental Setup}
\label{sec:experiment-setup}
\paragraph{Datasets and Metrics.} Following~\citet{deng2023implicit}, we employ the four-digit ($4 \times 4$) and five-digit ($5\times5$) multiplication problems from the BIG-bench benchmark~\citep{suzgun-etal-2023-challenging}, known to be challenging for LLMs to solve without CoT. Given that arithmetic reasoning is just one type of the reasoning ability, we also incorporate a boolean logical reasoning task~\citep{zhu2023promptbench}. For more complex tasks, grade school math problems require both language understanding and mathematical reasoning, so we adopt the widely-used GSM8K dataset~\citep{Cobbe2021TrainingVT}. We use the augmented training data from~\citet{deng2023implicit} and keep all original test sets unchanged. The statistics are listed in Appendix~\ref{app:tab:datasets}. For both datasets, we use accuracy to measure the exact match accuracy of predicting the final answer, and throughput to measure the number of samples processed per second (it/sec) during inference with a batch size of $1$.

\paragraph{Base Models.} When training from scratch, we follow DiffuSeq\footnote{\url{https://github.com/Shark-NLP/DiffuSeq}} to use a 12-layer Transformer~\citep{vaswani2017attention} encoder with similar size as GPT2-small (124M). We also use Plaid\footnote{\url{https://github.com/igul222/plaid}} (1.3B)~\citep{gulrajani2023likelihood}, SEDD-small\footnote{\url{https://github.com/louaaron/Score-Entropy-Discrete-Diffusion}} (170M) and SEED-medium (424M)~\citep{lou2023discrete} as pre-trained diffusion language models for further fine-tuning. Both Plaid and SEDD are pre-trained on OpenWebText~\citep{Gokaslan2019OpenWeb,pile}, which is similar to that in GPT2, and the pre-training perplexity of Plaid and SEDD-small is on par with GPT2-small.

\paragraph{Baselines.}
We consider \textbf{Answer-only} and \textbf{CoT} as reasoning paradigms for comparison. Another important baseline is \textbf{Implicit CoT}~\citep{deng2023implicit}, which distills thoughts into transformer layers to accelerate CoT reasoning. We use {GPT-2}~\citep{Brown2020LanguageMA} at various scales (i.e., small 124M, medium 355M, and large 774M) as model baselines, known as conventional autoregressive language models. We mainly consider fine-tuning the model due to the relatively small model size, but we also consider prompting the strong commercial LLM \textbf{ChatGPT} \texttt{gpt-3.5-turbo-1106} using CoT {few-shot} demonstrations for completeness. 
We use $5$-shot in the few-shot setting. 

\paragraph{Implementation Details.}
During tokenization, we treat all the digits as individual tokens. For \ourmodelmpdot, we append a special token <EOS> to the last thought, so when the model generates a thought followed by <EOS>, it stops generating further, which enables the model to decide the number of rationales dynamically. 
We conduct all the experiments on 8 NVIDIA V100-32G GPUs. During training, we set $\epsilon_{min}$ to be $0.95$ as we find decreasing the probability of oracle demonstration hinders model training. We choose coupled sampling $\gamma=0.01,k=1$ and self-consistency $m=20$. Following Plaid, we also adopt self-conditioning~\citep{chen2022analog} during training. 
During inference, we set both the temperature of the score and output logit to 0.5 to sharpen the predicted output distribution while maintaining the ability to generate diverse samples. The sampling timesteps $T$ is dynamic. By default, we set it to be $64$. Considering that simple tasks do not necessitate an excessively large number of steps, we opt to reduce $T$ while ensuring there is no notable performance drop. Other details are in Appendix~\ref{sec:app-imp}.

\begin{table*}[t]
\setlength{\tabcolsep}{3.5pt}
\centering
\caption{The main results on different problem-solving reasoning tasks. \textbf{Acc} ($\uparrow$) is to measure the exact match accuracy of the predicted final answer. \textbf{Throughput} ($\uparrow$) measures the number of samples processed per second during test with batch size equals to $1$. The baseline results for Mult. and GSM8K datasets are taken from the implicit CoT paper~\citep{deng2023implicit} and have been validated for reproducibility by us. Bracketed numbers indicate the self-consistency results.}
\vskip 0.15in
\scalebox{0.85}{
\begin{tabular}{llclclclc}
\toprule
\multicolumn{1}{c}{\multirow{2}{*}{Models}} & \multicolumn{2}{c}{$4\times4$ Mult.} & \multicolumn{2}{c}{$5\times5$ Mult.} & \multicolumn{2}{c}{Boolean logic} & \multicolumn{2}{c}{GSM8K-Aug} \\
\cmidrule(lr){2-3}\cmidrule(lr){4-5}\cmidrule(lr){6-7} \cmidrule(lr){8-9}
\multicolumn{1}{c}{}                                                            & Acc     & \small{Throughput}    & Acc     & \small{Throughput}    & Acc           & \small{Throughput} & Acc           & \small{Throughput}    \\
\midrule
Answer-only & & & & & & & &  \\
\quad GPT2-small                                           & 28.7     & 13.2          & 1.2    & 11.1      & 98.8 &  16.2  & 13.3          & 24.7           \\
\quad GPT2-medium                                           & 76.2     & 7.0           & 1.9    & 5.9       & 100 & 9.6 & 17.0          & 9.1           \\
\quad GPT2-large                                            & 33.6     & 4.8           & 0.9     & 4.0         & 100 & 7.4  & 12.7          & 9.1           \\
\quad ChatGPT (few-shot)                                                       & 2.2    & 1.0           & 0.0     & 1.4        & 67.6 & 0.5  & 28.1          & 1.8          \\

\midrule
Chain-of-Thoughts (CoT) & & & & & & & &\\
\quad GPT2-small                                              & 100     & 2.3           & 100     & 1.5       & 100 &  0.8  & 39.0 \small{(41.6)}          & 2.0           \\
\quad GPT2-medium                                             & 100     & 1.2           & 100     & 0.8       & 100 & 0.5   & 43.9          & 1.1           \\
\quad GPT2-large                                             & 100     & 0.8           & 99.3     & 0.6         & 100 & 0.3  & 44.8          & 0.7           \\
\quad ChatGPT (few-shot)                                                            & 42.8    & 0.1           & 4.5     & 0.1         & 75.8 &  0.2 & 61.5          & 0.2          \\
\midrule
Implicit CoT  & & & & & & \\
\quad GPT2-small                                                              & 96.6    & 8.9           & 9.5     & 7.9       & - & -   & 20.0          & 16.4          \\
\quad GPT2-medium                                                              & 96.1    & 4.8           & 96.4    & 4.3       & - & -   & 21.9          & 8.7           \\

\midrule
\textbf{Diffusion-of-Thoughts (DoT)} & & & & & & & & \\
\quad From-scratch   & 100    & 62.5       & 100     & 61.8     & 100 &  55.2  & 4.6        & 22.7        \\
\quad Plaid    & 100     &    24.3       & 100     &     21.3   & 100 & 10.2  & 32.6 \small{(36.3)}    &    0.3      \\
\quad SEDD-small    &   100  &  59.2    & 100    &   55.5  & 100 &  33.3  &  45.3 \small{(51.8)}   &    1.0    \\

\quad SEDD-medium    &   100  &   31.8     &   100  &   28.5    & 100 &  17.2  &  53.5 \small{(59.4)}   &      0.5  \\

\midrule
\textbf{Diffusion-of-Thoughts (\ourmodelmpdot)} & & & & & & & &\\
\quad From-scratch   & 100    & 11.8       & 100     & 9.5      & 100 & 3.7  & 5.5         & 8.6        \\
\quad Plaid    & 100     &    4.3     & 100   &     3.9   & 100 & 1.0  & 37.7    &    0.1      \\
\quad SEDD-small   &  100  &  9.9   & 100   &   9.2  & 100 &  3.3 &  43.2 &   0.2    \\
\quad SEDD-medium   &  100  &   4.5     &  100  &  4.0   & 100 &  1.7  &  53.3 &    0.1   \\

\bottomrule
\end{tabular}}
\label{tab:main-table}
\end{table*}

\subsection{Results on Digit Multiplication and Boolean Logic}
\label{sec:exp-main-digit}

We first train DoT for digit multiplication tasks and a boolean logical reasoning task as the preliminary investigation, as shown in the left part of Table~\ref{tab:main-table}. We observe that neither ChatGPT nor the distilled Implicit CoT model can reach 100\% accuracy. GPT-2 can be fine-tuned to achieve high accuracy but sacrifices throughput during CoT. Interestingly, DoT can attain 100\% accuracy for these tasks while maintaining significant throughput with diffusion sampling steps set at $1$ for multiplication datasets and $2$ for the boolean logical dataset, achieving maximum 27$\times$ speed-up compared to GPT-2. This preliminary finding indicates that DoT performs well in modeling exact math computation or boolean logic reasoning and benefits from its computational efficiency.

\subsection{Results on Grade School Math}
\label{sec:exp-main-gsm}

We now move on to a much more complex grade school math task GSM8K as shown in the right part of Table~\ref{tab:main-table}. We first consider training DoT from scratch as in the previous tasks, but we are only able to achieve an accuracy of around $5\%$, which is much lower than the fine-tuned version of GPT-2. This indicates the pre-trained natural language understanding capability is vital for grade school math. 
Once DoT is extended based on the pre-trained diffusion language models Plaid and SEDD, the performance is significantly improved after fine-tuning, where the DoT based on SEDD-medium outperforms similar-sized GPT2-medium with CoT by around 10\%. 
Additionally, multi-pass DoT, with casual bias, performs slightly better than single-pass one on Plaid, while the latter is more efficient. 
 The performance gap between SEDD and Plaid also highlights the importance of the training objective in pretraining diffusion LMs.
Finally, we find that self-consistency further yields substantial improvements in DoT models owing to the diverse generations of diffusion model (\S\ref{sec:exp-self-consis}).

\begin{figure}[t]
  \centering
\begin{minipage}{.48\linewidth}
\centering
\begin{tabular}{lc}
\toprule
Models                                                          &   Acc. (\%) $\uparrow$          \\
\midrule
Continue pre-training  &   0.5  \\
DoT-finetune&  32.6    \\
\quad (-) scheduled sampling &  31.2   \\
DoT$^{\text{MP}}$-finetune  & 37.7        \\
\quad (-) coupled sampling &  35.5    \\
\bottomrule
\end{tabular}
\vspace{20pt}
\captionof{table}{Ablation of Plaid DoT on GSM8K. 
} 
\label{tab:ablation-gsm}
\end{minipage}
\begin{minipage}{.45\linewidth}
\centering
\includegraphics[width=0.73\textwidth]{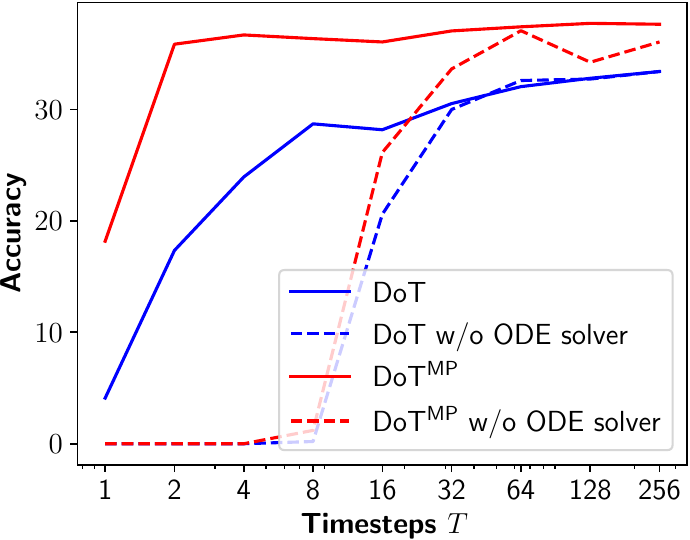}
\vspace{-5.5pt}
\caption{The effectiveness of ODE solver in speedup inference of Plaid DoT.}
\vspace{10pt}
\label{fig:timestep-t}
\end{minipage}

\vspace{-16pt}
\end{figure} 

We further explore several alternatives and conduct an ablation study as in Table~\ref{tab:ablation-gsm} when fine-tuning Plaid. As discussed above, continuing pre-training Plaid using the GSM8K-augmented dataset and performing reasoning with gradient-based conditioning is not a good choice for fine-tuning diffusion LMs on downstream tasks, because reasoning tasks require more specific guidance. An example of groundtruth and recovered text is shown below, where bold words in the query part are incorrectly recovered:
\begin{tcolorbox}[breakable]
\textit{Groundtruth}: Two trains leave San Rafael at the same time. They begin traveling westward, both traveling for 80 miles. The next day, they travel northwards, covering 150 miles. What's the distance covered by each train in the two days? <<2$*$80=160>> <<150$*$2=300>> <<300+160=460>> <<460/2=230>> \#\#\#\# 230
\\

\textit{Recovered Text}: \textbf{Three} trains leave San Juan at the same time. They \textbf{start} traveling westward, both traveling for 80 miles. The next day, they travel \textbf{southward}, covering 150 miles. What's the distance covered by each train in the two days? <<3$*$80=180>> <<180+80+150=340>> <<340/30=12.5>> \#\#\#\# 12.5
\end{tcolorbox}
We can see there are three recovered query tokens that exhibit minor differences due to soft gradient guidance, causing interference with the model's comprehension of the problem. 
The ablation of two sampling strategies proposed in \S\ref{sec:training} showcases their effectiveness. This provides evidence that better denoising models are trained using our training-time sampling strategies, allowing DoT models to self-correct more effectively during inference. Further analysis about self-correction is listed in \S\ref{sec:exp-self-corr}. In Figure~\ref{fig:timestep-t}, we further show the conditional ODE solver substantially speeds up the inference of continuous diffusion model Plaid, ensuring a decent performance with only 8 generation timesteps.

\begin{figure}[t]
\centering
\includegraphics[width=0.95\linewidth]{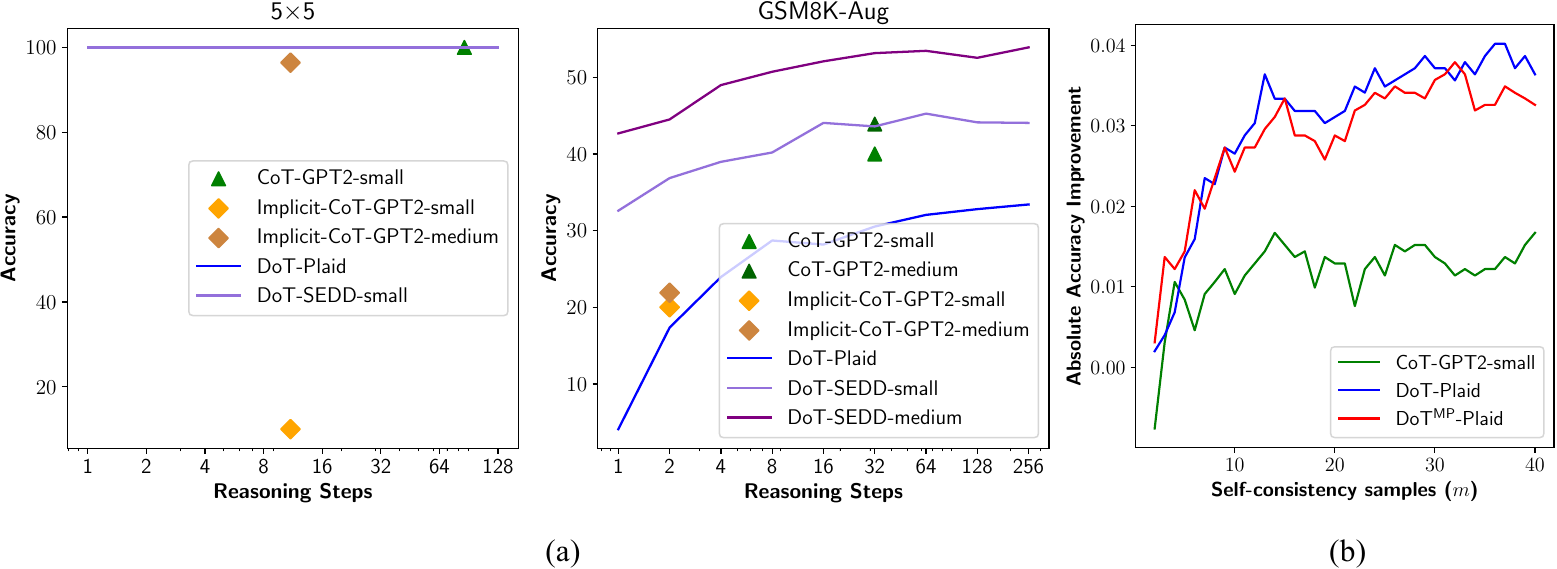}
\caption{
\textbf{(a)} Accuracy over reasoning steps using various methods. We measure the reasoning steps as the average number of calling the model \texttt{forward} function for instances from the test set. DoT provides a flexible way to balance accuracy and efficiency through the reasoning steps. \textbf{(b)} Absolute accuracy improvement versus samples in self-consistency per instance on the GSM8K dataset with Plaid DoT. 
}
\label{fig:reasoning-efficiency}
\end{figure}

\subsection{Reasonability-efficiency Trade-off}
\label{sec:exp-trade-off}
The community has devoted substantial efforts to improve the reasoning capabilities of left-to-right language models, such as refining instructions~\citep{Kojima2022LargeLM,Zhou2022LargeLM}, finding better demonstrations~\citep{fu2022complexity,Wu2022SelfAdaptiveIL,Ye2023CompositionalEF}, and designing elaborate decoding algorithm~\citep{shinn2023reflexion,xie2023decomposition,yao2023tree}. Non-autoregressive diffusion models naturally provide another simple way to enhance reasoning by allocating more timesteps during inference, albeit at the expense of efficiency.
We show such efficiency trade-off in Figure~\ref{fig:reasoning-efficiency}(a), where we measure the reasoning steps as the average number of calling the model \texttt{forward} function for all the instances from the test set. For CoT and Implicit CoT baselines, we treat reasoning steps as the average number of output tokens for all the test instances\footnote{
Here we define reasoning steps of Implicit CoT as the times of forwarding the whole model instead of the layers of transformers,  considering that the former reflects the inference speed.}. 

Given a small budget of reasoning steps (e.g., 1 or 2) on simpler tasks such as 5$\times$5, both DoT-Plaid and DoT-SEDD already have an accuracy of 100\%, and no more reasoning steps are needed. For such cases of simple tasks, only a little computation cost is required for our method. 
For complex tasks such as GSM8K, we find DoT performance can continuously improve by allowing more reasoning steps, which indicates DoT can be efficient
if we can sacrifice performance in certain scenarios. Specifically, DoT-SEDD-medium outperforms autoregressive CoT-GPT2-medium when we allocate 32 generation timesteps, and the performance continues improving when we increase the timesteps.
In comparison, CoT and Implicit CoT with the autoregressive model are hard to be more efficient given their nature of token-by-token prediction.
Overall, with DoT, we can flexibly control the trade-off between efficiency and performance for tasks with different difficulty levels.

\subsection{Self-consistency in DoT}
\label{sec:exp-self-consis}
Figure~\ref{fig:reasoning-efficiency}(b) shows the effectiveness of the self-consistency mechanism for Plaid DoT and its variant. We can see self-consistency improves both DoT and \ourmodelmpdot, which is in line with the effectiveness of self-consistency for auto-regressive models~\citep{wang2022self}.
From Table~\ref{tab:main-table}, SEDD DoT is also significantly improved by self-consistency. This benefits from the diversity generation in DoT.
We observe that DoT can generate diverse reasoning paths, such as \texttt{<3*3=9><9*60=540>} and \texttt{<3*60=180><180*3=540>} for the same question, providing cross-validation when selecting the most ``consistent'' answer.
Note that different from autoregressive models, where diversity usually relies on decoding algorithms~\citep{Fan2018HierarchicalNS,holtzman2019curious}, the natural advantage of the diffusion models is to generate different sentences with different random noises at each timestep.

\subsection{Self-correction in DoT}
\label{sec:exp-self-corr}

In this section, we provide several cases in Table~\ref{fig:case-self-correct} to show the self-correction ability of Plaid DoT, which acts as a distinct difference between diffusion models and autoregressive models. In the first case, we can see the model figures out all the correct thoughts together with only a single reasoning step (i.e., a single calling of the model \texttt{forward} function), and obtains the correct final answer in the second step. This mirrors how humans think in both fast and slow modes~\citep{Kahneman2011ThinkingFA}. In the second case where the problem is slightly harder, the model cannot give concrete thoughts in the first step but can still produce the correct answer through the later ``slow'' thinking process. We can see the solution framework, roughly outlining how the task will be carried out, is established at the very beginning, and then the subsequent work is for refining and improving, which is also similar to how human performs a complex task.
Interestingly, in DoT, the correct thoughts may not appear in a left-to-right paradigm as in the traditional chain-of-thought process. The third case serves as compelling evidence to illustrate this distinctive nature of diffusion-of-thought and how it diverges from the chain-of-thought approach. In step 4 the model has a wrong intermediate thought \texttt{<2*3=4>} with the latter thoughts and final answer computed correctly first. In the next step, the error in the wrong intermediate thought is fixed, which suggests both prior and latter thoughts can help in the prediction of the current thought. Furthermore, from these three cases, we observed that the model tends to maintain its prediction after it considers the answer to be complete. This suggests we can further enhance the inference efficiency by incorporating mechanisms such as early exit~\citep{Graves2016AdaptiveCT}, and easier tasks can get earlier exits as observed in Table~\ref{fig:case-self-correct}.

\begin{table*}
\caption{Cases that show the predictions of Plaid DoT at each time-step $t$ with $T$=8 on the GSM8K test set.
The incorrect thoughts are marked in bold red and we omit some correct predictions when $t<4$. 
The difficulty level of the questions increases from left to right. 
}
\centering
\vskip 0.15in
\includegraphics[width=0.99\textwidth]{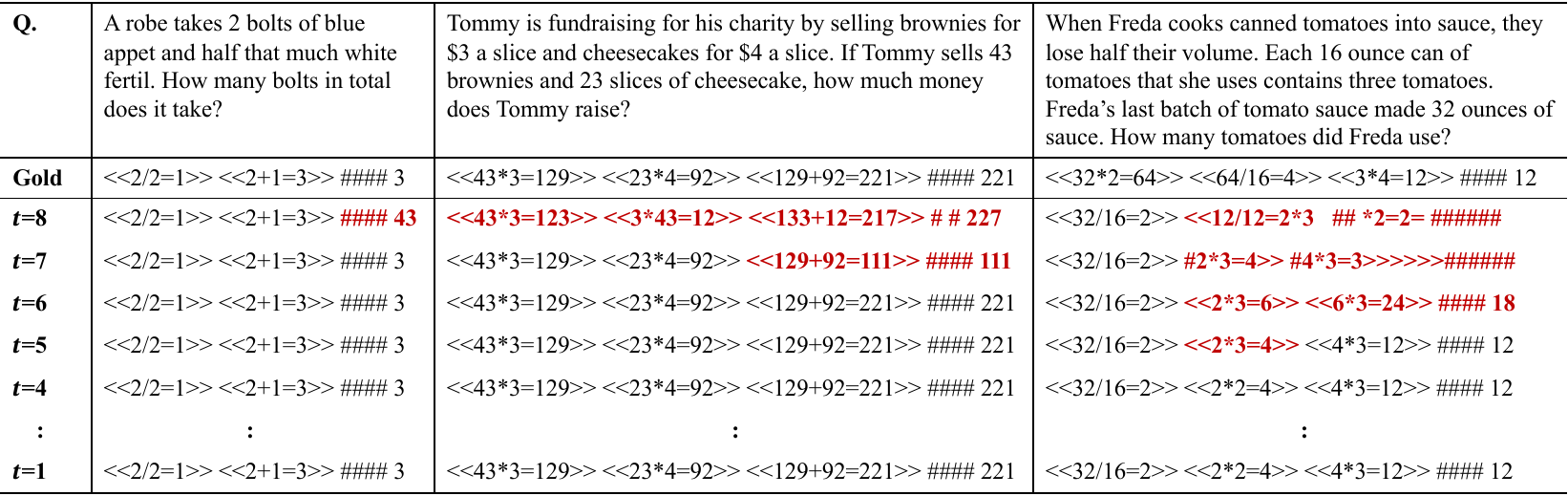}
\label{fig:case-self-correct}
\end{table*}
\section{Related Work}

\subsection{Diffusion Models for Text}
Building upon advancements in diffusion models for image generation~\citep{ho2020denoising, song2020denoising}, text continuous diffusion~\cite{li2022diffusion,gong2022diffuseq} employs an embedding function to transform discrete text into the continuous space. Besides, discrete diffusion models~\citep{hoogeboom2021argmax,austin2021structured} directly introduce discrete noise to accommodate the discrete nature of texts, demonstrating significant potential~\citep{Zheng2023ARD,lou2023discrete}.
Numerous studies have shown that diffusion models can efficiently generate diverse texts~\citep{gong-etal-2023-diffuseq, gao2022difformer}, and achieve competitive performance in various sequence-to-sequence NLP tasks, including machine translation~\citep{Yuan2022SeqDiffuSeqTD,ye2023dinoiser}, summarization~\citep{Zhang2023DiffuSumGE}, code generation~\citep{singh-etal-2023-codefusion}, and style transfer~\citep{horvitz2023paraguide}. 
In this work, we explore diffusion model for mathematical reasoning tasks.

\subsection{Pre-train and fine-tune Diffusion LMs}
The pre-training and fine-tuning paradigm, while a familiar concept in NLP before the era of prompting methods~\citep{liu2023pre}, remains relatively under-explored for diffusion language models. Prior efforts include initializing diffusion models with pre-trained masked language models such as BERT~\citep{he2022diffusionbert} and RoBERTa~\citep{Zhou2023DiffusionNATSD} and XLM-RoBERTa~\citep{ye2023diffusion}. GENIE~\citep{lin2023text} adopts paragraph denoising to train encoder-decoder models, proving beneficial for summarization tasks. Plaid~\citep{gulrajani2023likelihood} and SEDD~\citep{lou2023discrete} are pioneers in pre-train diffusion language models from scratch, attaining comparative or better perplexity scores over GPT-2~\citep{Brown2020LanguageMA}. To the best of our knowledge, we are the first to explore the fine-tuning of a pre-trained diffusion language model for reasoning tasks. 

\subsection{Reasoning Paradigms}
Large language models usually excel in performing system-1~\citep{Stanovich2000IndividualDI} tasks that are processed quickly and intuitively by humans but struggle in system-2 tasks, which require deliberate thinking~\citep{Brown2020LanguageMA,Wei2022ChainOT,Suzgun2022ChallengingBT}. The chain-of-thought reasoning paradigm~\citep{Nye2021ShowYW,Wei2022ChainOT,Kojima2022LargeLM} has been widely employed to elicit reasoning abilities and can be further improved with various techniques. For instance, self-consistency~\citep{wang2022self}
samples a diverse set of reasoning paths and selects the most consistent answer, while tree-of-thought~\citep{yao2023tree}
achieves different reasoning paths by tree search. Despite these advancements, errors introduced in intermediate CoT steps can lead to inaccurate answers~\citep{lanham2023measuring}, posing difficulties in self-correction~\citep{huang2023large}. Moreover, there are concerns about the inefficiency of CoT~\citep{deng2023implicit}. From the architecture perspective, we explore diffusion model as an alternative paradigm for reasoning.

\section{Conclusion and Limitation}
\label{sec:conclusion}

In this work, we propose diffusion-of-thought (DoT), integrating CoT reasoning with continuous diffusion models. We thoroughly evaluate DoT on representative mathematical reasoning tasks in various aspects, including their flexible control of reasoning efficiency, self-correction capability, and the ability to generate diverse reasoning paths. Considering pre-trained diffusion models are still in their early stages, particularly in terms of model scales compared to the more extensively studied autoregressive language models, our study presents an initial exploration into the reasoning ability of current diffusion language models. 
A notable limitation of DoT is its requirement for additional training to achieve accurate reasoning. With more powerful pre-trained diffusion models, we anticipate DoT can attain comparative or better generalization capabilities of auto-regressive language models while removing the need for specialized training. Moreover, extending the standard Transformer to other variants~\citep{gu2023mamba} is also a viable direction to further improve inference efficiency. 
Besides, the diffusion training techniques employed in this work are general and applicable to other tasks beyond mathematical reasoning. Extending our training recipes of diffusion language models to further scaled setups such as multi-task instruction tuning and other modalities~\citep{zhang2023multimodal,harvey2023visual}, is an interesting avenue for future research. 

\medskip
{
\small
\bibliography{reference_clean}
}

\newpage
\appendix
\onecolumn

\section{Derivations}
\label{sec:app-eq}
\subsection{Seq2Seq Modeling in DiffuSeq}
\label{sec:app-diffuseq}
To implement the diffusion model in seq2seq generation, we inherit the design from \textbf{DiffuSeq} \citep{gong-etal-2023-diffuseq}, which systematically defines the $forward~noising$ process and $reverse~denoising$ process on latent continuous space \textbf{z} as two major components of the model. 

\paragraph{Latent space configuration \textbf{z}.}
Following \citet{li2022diffusion}, \textbf{z} is constructed from an embedding function $\textsc{Emb}(\mathbf{w^z})$, which takes the discrete text $\textbf{w}^z$ as input. Particulatly, in Diffuseq \citep{gong-etal-2023-diffuseq}, $\textbf{w}^z$ contains $\mathbf{w}^x$ and $\mathbf{w}^y$ where $\mathbf{w}^x$ is the source sequence and $\mathbf{w}^y$ is the target sequence. The relationship is defined as $\mathbf{w}^{z}=\mathbf{w}^{[x; y]}$. They denote $\mathbf{z}_{t} = \mathbf{x}_t \oplus \mathbf{y}_t$ to simplify the wordings, where $\mathbf{x}_t$ and $\mathbf{y}_t$ represent parts of $\mathbf{z}_{t}$ that belong to $\mathbf{w}^x$ and $\mathbf{w}^y$, respectively.

\paragraph{Forward diffusion process $q(\mathbf{z}_{t} \vert \mathbf{z}_{t-1})$ \textbf{and}  $q(\mathbf{z}_{t} \vert \mathbf{z}_{0})$.}
The process of $forward~noising$ is to fractionally disrupt the content of input data $\mathbf{z}_0$, introduced as \textbf{partial noising} by \citet{gong2022diffuseq}. It is achieved by only applying Gaussian noise to $\mathbf{y}_t$ and preserving $\mathbf{x}_t$ with a masking scheme, denoted as $\mathbf{z}_{t} = [\mathbf{x}_t ; \mathbf{y}_t]$ with mask $[\mathbf{0};\mathbf{1}]$.

After the process of $forward~noising$ where $T$-step forward random disturbance is applied, the $\mathbf{z}_0$ is finally transformed into the partial Gaussian noise with $\mathbf{y}_T\sim \mathcal{N}(0, \mathbf{I}) $.
\begin{equation}
    q(\mathbf{z}_{t} \vert \mathbf{z}_{t-1}) = \mathcal{N}(\mathbf{z}_{t};\sqrt{1-\beta_t}\mathbf{z}_{t-1}, {\beta}_t \mathbf{I}),
\end{equation}
\begin{equation}
    q(\mathbf{z}_{1:T} \vert \mathbf{z}_0) = \prod^T_{t=1} q(\mathbf{z}_t \vert \mathbf{z}_{t-1})
\end{equation}
where $t = 1, 2,...,T$ and $\{\beta_t \in (0,1)\}_{t=1}^T$ are the variance schedule. A reparameterization trick could be applied to the above process to attain a closed-form representation of sampling $\mathbf{z}_t$ at any arbitrary time step $t$. Let $\alpha_t=1-\beta_t$ and $\bar{\alpha}_t = \prod_{i=1}^t \alpha_i$, the equation is reduced to:

\begin{equation}
\begin{aligned}
\label{eq:zt-d}
    \mathbf{z}_t = &\sqrt{\alpha_t} \mathbf{z}_{t-1}+\sqrt{1-\alpha_t}\epsilon_{t-1}=\sqrt{\alpha_t\alpha_{t-1}} \mathbf{z}_{t-2}+\sqrt{1-\alpha_t\alpha_{t-1}}\bar{\epsilon}_{t-2}\\
    =&...=\sqrt{\bar{\alpha_t}}\mathbf{z}_0+\sqrt{1-\bar{\alpha}_t}\epsilon,
\end{aligned}
\end{equation}

where $\boldsymbol{\epsilon}_{t-1}, \boldsymbol{\epsilon}_{t-2}, \dots \sim \mathcal{N}(\mathbf{0}, \mathbf{I})$ and ${\boldsymbol{\epsilon}}$ merges all the Gaussians. In the end:
\begin{equation}
    q(\mathbf{z}_t \vert \mathbf{z}_0) 
    = \mathcal{N}(\mathbf{z}_t; \sqrt{\bar{\alpha}_t} \mathbf{z}_0, (1 - \bar{\alpha}_t)\mathbf{I})
\end{equation}
A $sqrt$ noise schedule is applied according to the Diffusion-LM~\citep{li2022diffusion}, that is, $\bar{\alpha}_t=1-\sqrt{t/T+s}$ with $s$ as a small constant at the start of the noise level.

\paragraph{Posterior $q(\mathbf{z}_{t-1}|\mathbf{z}_t,\mathbf{z}_0)$.} Derived by Bayes’ rule, the posterior is given by:
\begin{equation}
\begin{aligned}
    q(\mathbf{z}_{t-1} \vert \mathbf{z}_t, \mathbf{z}_0) 
    &= q(\mathbf{z}_t \vert \mathbf{z}_{t-1}, \mathbf{z}_0) \frac{ q(\mathbf{z}_{t-1} \vert \mathbf{z}_0) }{ q(\mathbf{z}_t \vert \mathbf{z}_0) }
\end{aligned}
\end{equation}

Given the above relationship, the posterior is still in Gaussian form. After applying the Eq.~(\ref{eq:zt-d}) to it, the mean of  $q(\mathbf{z}_{t-1}|\mathbf{z}_t,\mathbf{z}_0)$ could be derived:
\begin{equation}
\label{eq:ut}
    \mu_t(\mathbf{z}_t,\mathbf{z}_0)=\frac{\sqrt{\alpha_t}(1-\bar{\alpha}_{t-1})}{1-\bar{\alpha}_t}\mathbf{z}_t+\frac{\sqrt{\bar{\alpha}_{t-1}}(1-\alpha_t)}{1-\bar{\alpha}_t}\mathbf{z}_0,
\end{equation}

\paragraph{Backward generative process $p_\theta(\mathbf{z}_{0:T}|\mathbf{z}_T)$.} After the $forward~noising$ process is defined and the training is completed, the $reverse~denoising$ process then denoises $\mathbf{z}_t$, aiming to recover original $\mathbf{z}_0$ with the trained Diffuseq model $\z_{\theta}(\mathbf{z}_t, t)$. This process is defined as:
\begin{equation}
\label{diffu_rev}
    p_{\theta}(\mathbf{z}_{0:T})=p_\theta(\mathbf{z}_T)\prod_{t=1}^Tp_{\theta}(\mathbf{z}_{t-1}|\mathbf{z}_t)
\end{equation}

\begin{equation}
\label{diffu_one_step_learning}
    p_{\theta}(\mathbf{z}_{t-1}|\mathbf{z}_t)=\mathcal{N}(\mathbf{z}_{t-1};\mu_{\theta}(\mathbf{z}_t, t), \sigma_{\theta}(\mathbf{z}_t, t)),
\end{equation}
and the initial state $p_\theta(\mathbf{z}_T)$ is defined as $\mathcal{N}(0,\mathbf{I})$.
\paragraph{Training objective $\mathcal{L}_{\text{VLB}}$.}
Inherited from Diffuseq \citep{gong-etal-2023-diffuseq}, the training objective is to recover the original $\mathbf{z}_0$ by denoising $\mathbf{z}_t$ as in Eq.~(\ref{diffu_rev}). The learning process as Eq.~(\ref{diffu_one_step_learning}) is modeled by Diffuseq: $\z_{\theta}(\mathbf{z}_t, t)$, where the $\mu_{\theta}(\cdot)$ and $\sigma_{\theta}(\cdot)$ serve as the parameterization of the predicted mean and standard {deviation} of $ q(\mathbf{z}_{t-1}|\mathbf{z}_{t})$ in the $forward~noising$ process respectively. The input $\mathbf{x}_t$ serves as the condition during the $reverse~denoising$ process as the partial noising is adopted in the $forward~noising$.

Typically, a transformer architecture is adopted to model $\z_{\theta}$, which is capable of modeling the semantic relation between $\mathbf{x}_t$ and $\mathbf{y}_t$ instinctively. The variational lower bound ($\mathcal{L}_\text{VLB}$) is computed as follows:

\begin{equation}
\label{eq:vlbloss}
\begin{aligned}
    \mathcal{L}_{\text{VLB}}(\mathbf{w}^z) & = \mathbb{E}_{q({\mathbf{z}_0}\mid \mathbf{w}^z)} \Bigg[ \underbrace{\log\frac{ q(\mathbf{z}_T|\mathbf{w}^z)}{p_{\theta}(\mathbf{z}_T)}}_{\text{Prior loss}}
    + \underbrace{\textstyle \mathcal{L}_{\text{VLB}}(\mathbf{z}_0)}_{\text{Diffusion loss}}
     \underbrace{-\log p_\theta(\mathbf{w}^z|\mathbf{z}_0)}_{\text{Rounding loss}}\Bigg],
\end{aligned}
\end{equation}
where the diffusion loss is the same as the continuous diffusion loss in DDPM~\citep{ho2020denoising}, which is given by:
\begin{equation}
    \mathcal{L}_{\text{VLB}}(\mathbf{z}_0) =\mathbb{E}_{ q(\mathbf{z}_{1:T}|\mathbf{z}_0)}
    \Bigg[
    \underbrace{\log\frac{ q(\mathbf{z}_T|\mathbf{z}_0)}{p_{\theta}(\mathbf{z}_T)}}_{\mathcal{L}_T}  
    + \underbrace{\sum_{t=2}^T \log{\frac{ q(\mathbf{z}_{t-1}|\mathbf{z}_0,\mathbf{z}_t)}{p_{\theta}(\mathbf{z}_{t-1}|\mathbf{z}_t)}}}_{\mathcal{L}_{T-1} + \dots + \mathcal{L}_1}
    - {\underbrace{\log{ p_{\theta}(\mathbf{z}_0|\mathbf{z}_1)}}_{\mathcal{L}_0}}
    \Bigg].
\end{equation}
Here the prior loss and $\mathcal{L}_T$ is considered as a constant when the noising schedule $q$ is fixed and $p_\theta(\mathbf{z}_{T})=\mathcal{N}(0, \mathbf{I})$. 

After reweighting each term (i.e., treating all the loss terms across time-steps equally) as in ~\citet{ho2020denoising} and using the Monte Carlo optimizer, the training objective can be further simplified as: 
\begin{equation}
\label{eq:loss}
\begin{aligned}
\min_{\theta}\; \mathcal{L}_{\text{VLB}}(\mathbf{w}^z) & \rightarrow  \min_{\theta}\mathbb{E}_{q(\mathbf{z}_{0:T}\mid \mathbf{w}^z)}\left[
\sum_{t=2}^T||\mathbf{z}_0-\z_{\theta}(\mathbf{z}_t, t)||^2 + ||\textsc{Emb}(\mathbf{w}^{z})-\z_{\theta}(\mathbf{z}_1, 1)||^2-\log p_{\theta}(\mathbf{w}^{z}|\mathbf{z}_0)\right] \\
& \rightarrow \min_{\theta}\left[ \sum_{t=2}^T||\mathbf{y}_0-\tilde \z_{\theta}(\mathbf{z}_t, t)||^2 + {||\textsc{Emb}(\mathbf{w}^y)-\tilde \z_{\theta}(\mathbf{z}_1, 1)||^2} + \mathcal{R}(||\mathbf{y}_0||^2)\right]\\
& \rightarrow \min_{\theta}\left[ \sum_{t=1}^T||\mathbf{y}_0-\tilde \z_{\theta}(\mathbf{z}_t, t)||^2+ \mathcal{R}(||\mathbf{y}_0||^2)\right],
\end{aligned}
\end{equation}
where $\tilde \z_{\theta}(\mathbf{z}_t, t)$ is used to denote the fractions of recovered $\mathbf{z}_0$ corresponding to $\mathbf{y}_0$. $\mathcal{R}(||\mathbf{y}_0||^2)$) is the regularization term which regularizes the embedding learning. The embedding function is shared between source and target sequences, contributing to the joint training process of two different feature spaces.

\subsection{Pre-trained Plaid}
The Plaid model~\citep{gulrajani2023likelihood} mostly adopts the variational diffusion model (VDM) framework~\citep{kingma2021variational} and we illustrate its forward, reverse, and loss calculations in this section. When fine-tuning Plaid 1B, we use the VDM formulation and apply the same sequence-to-sequence modification as in DiffuSeq. This involves imposing partial noise on $\mathbf{z}_t$ and keeping the source condition sentence anchored as un-noised.
\paragraph{Forward diffusion process $q(\mathbf{z}_t|\mathbf{z}_0)$ and $q(\mathbf{z}_t|\mathbf{z}_s)$.} 
The distribution of latent $\mathbf{z}_t$ conditioned on $\mathbf{z}_0$ is given by:
\begin{equation}
\label{eq:zt}
    q(\mathbf{z}_t \vert \mathbf{z}_0) = \mathcal{N}(\alpha_t \mathbf{z}_0, \sigma_t^2\mathbf{I}).
\end{equation}
After reparameterization, we have $\mathbf{z}_0=(\mathbf{z}_s-\mathbf{\epsilon}_1\sigma_s)/\alpha_s$
and $\mathbf{z}_t=(\alpha_t/\alpha_s)\mathbf{z}_s - (\alpha_t\sigma_s/\alpha_s)\epsilon_1+\sigma_t\epsilon_2$, where $\mathbf{\epsilon}_1 \sim \mathcal{N}(0,\mathbf{I})$ and $\mathbf{\epsilon}_2 \sim \mathcal{N}(0,\mathbf{I})$. Then after merging two uniform Gaussians, the distribution of $\mathbf{z}_t$ given $\mathbf{z}_s$, for any $0 \leq s < t \leq 1$, is given by:
\begin{equation}
    q(\mathbf{z}_t|\mathbf{z}_s) = \mathcal{N}\left(\alpha_{t|s} \mathbf{z}_s, \sigma^2_{t|s} \mathbf{I}\right),
\label{eq:zt_given_zs}
\end{equation} 
where $\alpha_{t|s} = \alpha_t/\alpha_s$ and $\sigma_{t|s}^2 = \sigma_t^2 - \alpha_{t|s}^2\sigma_s^2$.
The variance-preserving special case gives $\alpha_t=\sqrt{1-\sigma_t^2}$. 
In VDM, the noise schedule $\alpha_t$ and $\sigma_t^2$, which specify how much noise to add at each time in the diffusion process, are parameterized as a scalar-to-scalar neural network $\boldsymbol{\eta}$ that satisfies $\sigma_t^2=\operatorname{sigmoid}\left(\gamma_{\boldsymbol{\eta}}(t)\right)$ and $\alpha_t^2=\operatorname{sigmoid}\left(-\gamma_{\boldsymbol{\eta}}(t)\right)$. This is different from previous practices that use a predefined function, e.g., DDPM~\citep{ho2020denoising} set the forward process variances to constants increasing linearly from $\beta_1=10^{-4}$ to $\beta_T=0.02$. 

\paragraph{Posterior $q(\mathbf{z}_s|\mathbf{z}_t,\mathbf{z}_0)$.}
The joint distribution of latent variables $(\mathbf{z}_s, \mathbf{z}_t, \mathbf{z}_u)$ at any subsequent timesteps $0 \leq s < t < u \leq 1$ is Markov: $q(\mathbf{z}_u | \mathbf{z}_t, \mathbf{z}_s) = q(\mathbf{z}_u | \mathbf{z}_t)$. Given the distributions above, we can verify through the Bayes rule that $q(\mathbf{z}_s|\mathbf{z}_t,\mathbf{z}_0)$, for any $0 \leq s < t \leq 1$, is also Gaussian given by: 
\begin{align}
q(\mathbf{z}_s|\mathbf{z}_t,\mathbf{z}_0) &= \mathcal{N}(\mathbf{\mu}_{Q}(\mathbf{z}_{t},\mathbf{z}_0; s,t), \sigma^2_Q(s,t) \mathbf{I})\\
\text{where\;\;}
\sigma^{2}_Q(s,t) &= \sigma^2_{t|s}\sigma^2_s / \sigma^2_t \label{eq:postvar}\\
\text{and\;\;}
\mathbf{\mu}_{Q}(\mathbf{z}_{t},\mathbf{z}_0; s,t) 
& = \frac{\alpha_{t|s}\sigma^{2}_s}{\sigma^{2}_t}\mathbf{z}_{t} + \frac{\alpha_s \sigma^{2}_{t|s}}{\sigma^{2}_{t}}\mathbf{z}_0. \label{eq:postmu}
\end{align}

\paragraph{Backward generative process $p_\theta(\mathbf{z}_s|\mathbf{z}_t)$.} 
In VDM, the reverse process or the generative process is also defined as a Gaussian that satisfies $p_\theta(\mathbf{z}_s|\mathbf{z}_t) = q(\mathbf{z}_s|\mathbf{z}_t, \mathbf{z}_0=\z_\theta(\mathbf{z}_t; t))$, i.e.\ the same as $q(\mathbf{z}_s|\mathbf{z}_t,\mathbf{z}_0)$, but with the original data $\mathbf{z}_0$ replaced by the output of the denoising model $\z_\theta(\mathbf{z}_t; t)$. Therefore, based on Eq.~(\ref{eq:postmu}), the mean of $p_\theta(\mathbf{z}_s|\mathbf{z}_t)$ is given by:
\begin{equation}
    \mathbf{\mu}_{\theta}(\mathbf{z}_{t}; s,t) = \frac{\alpha_{t|s}\sigma^{2}_s}{\sigma^{2}_t}\mathbf{z}_{t} + \frac{\alpha_s \sigma^{2}_{t|s}}{\sigma^{2}_{t}}\z_\theta(\mathbf{z}_t; t),
\end{equation}
and the variance is the same as Eq.~(\ref{eq:postvar}).

\paragraph{Continuous diffusion loss term $\mathcal{L}_{\text{VLB}}(\mathbf{z}_0)$.}
The prior loss and rounding loss in Eq.~(\ref{eq:vlbloss}) can be (stochastically and differentiably) estimated using standard techniques. We now derive an estimator for the diffusion loss in VDM. Different from Eq.(\ref{eq:loss}) which simplifies the loss term by reweighting, VDM adopts the standard loss formulation.
We begin with the derivations of diffusion loss for discrete-time diffusion with $t \in \{1,\dots,T\}$, which is given by:
\begin{align}
    \mathcal{L}_{\text{VLB}}(\mathbf{z}_0)
    &= 
    \sum_{t=1}^T
    \mathbb{E}_{q(\mathbf{z}_{t}|\mathbf{z}_0)} 
    D_{KL}[q(\mathbf{z}_{s}|\mathbf{z}_{t},\mathbf{z}_0)||p(\mathbf{z}_{s}|\mathbf{z}_{t})],
\end{align}
and we derive the expression of $D_{KL}(q(\mathbf{z}_s|\mathbf{z}_t,\mathbf{z}_0)||p_\theta(\mathbf{z}_s|\mathbf{z}_t))$ as follows:
\begin{align}
D_{KL}(q(\mathbf{z}_s|\mathbf{z}_t,\mathbf{z}_0)||p_\theta(\mathbf{z}_s|\mathbf{z}_t)) 
&= \frac{1}{2\sigma^2_Q(s,t)}
||\mathbf{\mu}_Q - \mathbf{\mu_\theta}||_2^2
\label{eq:KL_DPM}
\\
&= 
\frac{\sigma^2_t}{2\sigma^2_{t|s}\sigma^2_s}
\frac{\alpha_s^{2} \sigma^{4}_{t|s}}{\sigma^{4}_{t}}
||\mathbf{z}_0 - \z_\theta(\mathbf{z}_t; t)||_2^2\\
&= 
\frac{1}{2\sigma^2_s}
\frac{\alpha_s^{2} \sigma^{2}_{t|s}}{\sigma^{2}_{t}}
||\mathbf{z}_0 - \z_\theta(\mathbf{z}_t; t)||_2^2\\
&= 
\frac{1}{2\sigma^2_s}
\frac{\alpha_s^{2}(\sigma^{2}_t - \alpha_{t|s}^{2}\sigma^{2}_s)}{\sigma^{2}_{t}}
||\mathbf{z}_0 - \z_\theta(\mathbf{z}_t; t)||_2^2\\
&= 
\frac{1}{2}
\frac{\alpha_s^{2}\sigma^{2}_t/\sigma^2_s - \alpha_{t}^{2}}{\sigma^{2}_{t}}
||\mathbf{z}_0 - \z_\theta(\mathbf{z}_t; t)||_2^2\\
&= 
\frac{1}{2}\left(
\frac{\alpha_s^{2}}{\sigma^2_s} - \frac{\alpha_t^{2}}{\sigma^{2}_t}\right)
||\mathbf{z}_0 - \z_\theta(\mathbf{z}_t; t)||_2^2\\
&= 
\frac{1}{2}\left(
\text{SNR}(s) - \text{SNR}(t)\right)
||\mathbf{z}_0 - \z_\theta(\mathbf{z}_t; t)||_2^2,
\end{align}
where $\text{SNR}(t)=\alpha_t^2/\sigma_t^2$ and its physical meaning is signal-to-noise ratio.

After reparameterization of $\mathbf{z}_t$, the diffusion loss function becomes:
\begin{align}
\mathcal{L}_{\text{VLB}}(\mathbf{z}_0) &= \sum_{t=1}^T \mathbb{E}_{q(\mathbf{z}_t|\mathbf{z}_0)}[ D_{KL}(q(\mathbf{z}_s|\mathbf{z}_t,\mathbf{z}_0)||p_\theta(\mathbf{z}_s|\mathbf{z}_t))]\\
&= \frac{1}{2}\mathbb{E}_{\mathbf{\epsilon} \sim \mathcal{N}(0,\mathbf{I})}[\sum_{t=1}^T
\left(
\text{SNR}(s) - \text{SNR}(t)\right)
||\mathbf{z}_0 - \z_\theta(\mathbf{z}_t; t)||_2^2].
\end{align}
In practice, we follow Plaid to use the continuous-time diffusion formulation, where $t\in [0,1]$, and we can express $\mathcal{L}$ as a function of $\tau$ with $\tau \rightarrow 0$:
\begin{align}
\mathcal{L}_{\text{VLB}}(\mathbf{z}_0) =
\frac{1}{2}
\mathbb{E}_{\mathbf{\epsilon} \sim \mathcal{N}(0,\mathbf{I})}\int_{0}^{1}
\left[
\frac{\text{SNR}(t-\tau)-\text{SNR}(t)}{\tau}
||\mathbf{z}_0 - \z_\theta(\mathbf{z}_t;t) ||_2^2
\right]dt,
\end{align}
and let $\text{SNR}'(t)$ denote the derivative of the SNR function, this then gives:
\begin{align}
\mathcal{L}_{\text{VLB}}(\mathbf{z}_0) &= -\frac{1}{2}\mathbb{E}_{\mathbf{\epsilon}\sim\mathcal{N}(0,\mathbf{I})} \int_{0}^{1} \text{SNR}'(t) \left\rVert \mathbf{z}_0 - \z_\theta(\mathbf{z}_t;t) \right\lVert_{2}^{2} dt
.
\end{align}


\subsection{Pre-trained SEDD}
SEDD~\citep{lou2023discrete} is a discrete diffusion language model built based on discrete score matching~\citep{meng2022concrete}, which generalizes score matching~\citep{song2020denoising,song2020score} to the discrete data. We now denote $x$ as a categorical random variable and the following derivation can be extended to a sequence of variable $\x$ as well.

\paragraph{Concrete score.}
Instead of directly modeling $p_\theta(x)$ to approximate original data distribution $q(x)$, the core idea of discrete score matching is to learn a quantity known as the \textit{concrete score}~\citep{meng2022concrete} through a neural network:
\begin{equation} 
s_\theta(x)_y=\frac{p_\theta(y)}{p_\theta(x)} = \frac{e^{f_\theta(y)} / Z}{e^{f_\theta(x)} / Z} = \frac{e^{f_\theta(y)}}{e^{f_\theta(x)}},
\end{equation}
which eliminates normalizing constant $Z$ as in the energy-based model. In particular, this quantity is the categorical equivalent of the famous score function $\nabla_x \log p$ in continuous space. Regarding the choice of $y$, 
if we model the ratio for every possible $y$, we would have $V$ items given $V$ as the dimension of $x$, and $N^V$ items for $\x$ given $N$ as the sequence length of $\x$, which is computationally intractable. So we sparsify and only model "relevant" ratios based on whether ${y}$ is "close" to ${x}$. These relevant positions will be denoted as ${y} \sim {x}$, e.g., all sentences ${y}$ that differ from ${x}$ with Hamming distance 1.

\paragraph{Training objective.}
~\citet{lou2023discrete} define a learning objective named \textit{score entropy} to learn the neural network, which is given by:
\begin{equation} 
\mathbb{E}_{x \sim q} \left[\sum_{y \sim x} s_\theta(x)_y - \frac{q(y)}{q(x)} \log s_\theta(x)_y \right].
\end{equation}
Taking a derivative w.r.t. $s$ and setting it to 0, we see that this occurs when $s_\theta(x)_y = \frac{q(y)}{q(x)}$, which can be easily checked to be globally optimal as the function is convex as a function of $s$.
To handle the unknown term  $\frac{q(y)}{q(x)}$, they further propose \textit{denoising score entropy} motivated by \textit{denoising score matching}~\citep{song2020denoising} based on $q(x_t) = \sum_{x_0} q_{t\vert0}(x_t | x_0) q_0(x_0)$:
\begin{equation}
\label{eqn:dse} 
\mathbb{E}_{x_0 \sim q_0, t \sim U[0,T], x_t \sim q_{t\vert0}(x_t \vert x_0)} \left[\sum_{y \sim x_t} s_\theta(x_t,t)_{y} - \frac{q_{t\vert0}(y | x_0)}{q_{t\vert0}(x_t | x_0)} \log s_\theta(x_t,t)_{y}\right],
\end{equation}
where $q_{t\vert0}(\cdot \vert x_0)$ is a perturbation of a base density $q(\cdot)$ by a transition kernel, and the transition ratio $\frac{q_{t\vert0}(y \vert x_0)}{q_{t\vert0}(x_t \vert x_0)}$ is known by design.

\paragraph{Forward diffusion process.}
The transition $q_{t\vert0}(x_t \vert x_0)$ is a vector that represents a categorical distribution, and can be defined by a forward diffusion process $q_{t\vert0}(x_t \vert x_0)=\exp(\overline{\sigma}(t)Q)_{x_0}$, where $\overline{\sigma}(t)\in \mathbb{R}_{\geq 0}$ is the cumulative noise $\int_0^t \sigma(s) ds$ at timestep $t$ with a value close to 0 when $t$ is small and increasing when $t$ growing.
~\citet{lou2023discrete} use two standard transition matrices with special structures to implement matrix $Q$ following prior work \citep{austin2021structured}:
\begin{gather}
    Q^{\rm uniform} = \begin{bmatrix} 1 - V & 1 & \cdots & 1\\ 1 & 1 - V & \cdots & 1\\ \vdots & \vdots & \ddots & \vdots \\ 1 & 1 & \cdots & 1 - V\end{bmatrix}\\
    Q^{\rm absorb} = \begin{bmatrix} -1 & 0 & \cdots & 0 & 0\\ 0 & -1 & \cdots & 0 & 0\\ \vdots & \vdots & \ddots & \vdots & \vdots \\ 0 & 0 & \cdots & -1 & 0\\ 1 & 1 & \cdots & 1 & 0\end{bmatrix}
\end{gather}

One can view the above diffusion process by taking small $\Delta t$ Euler steps and randomly sampling the resulting transitions:
\begin{equation}\label{eqn:discrete_euler_base}
    q_{{t + \Delta t}\vert t}(x_{t + \Delta t} = y | x_t = x) \varpropto \delta_{xy} + Q_t(y, x) \Delta t + O(\Delta t),
\end{equation}
where 
 $Q_t = \sigma(t) Q$, and $O(\Delta t)$ represents terms that tend to zero at a faster rate than $\Delta t$.

\paragraph{Backward generative process.}
To simulate the diffusion defined above, one can use the Euler strategy to derive the time reversal of the forward process:
\begin{equation}
    q_{{t-\Delta t}\vert t}(x_{t-\Delta t} = y \mid  x_t = x) \varpropto \delta_{xy} + R_t(y, x) \Delta t + O(\Delta t),
\end{equation}
where $R_t$ is the reverse transition rate matrix that can be derived using Bayes rule: $R_t(y, x) = \frac{q_t(y)}{q_t(x)}Q_t(x,y)$. Each column of $R_t$ represents the transition probability from a token at timestep $t$ to other tokens at timestep $t-\Delta t$.
Let $p_\theta(x_{t-\Delta t} = y \mid  x_t = x)  = q_{{t-\Delta t}\vert t}(x_{t-\Delta t} = y \mid  x_t = x)$, we have:
\begin{equation}
    p_\theta(x_{t-\Delta t} = y \mid  x_t = x) \varpropto \delta_{xy} + R^\theta_t(y, x) \Delta t + O(\Delta t),
\end{equation}
where $R^\theta_t(y, x)=\sum_{x_0}q_{0\vert t}(x_0\vert x)\frac{q_{t\mid0}(y\mid x_0)}{q_{t\mid0}(x\mid x_0)}Q_t(x,y)=s_\theta(x,t)_{y}Q_t(x,y)$.
For a sequence of random variables $\mathbf{x}$, this is inefficient because only one position is modified per step. A natural alternative has been to use \textit{$\tau$-leaping} \citep{Gillespie2001ApproximateAS}, which performs an Euler step at each position simultaneously.

\subsection{Conditional ODE solver}
\label{app:conditional-ode-solver}

The sampling of continuous diffusion models can be implemented by solving the diffusion ODEs~\citep{song2020denoising,song2020score}. Specifically, sampling by diffusion ODEs needs to discretize the following ODE~\citep{song2020score} with $t$ changing from $T$ to $0$:
\begin{equation}
\label{eq:diffusion_ode_eps}
    \frac{\dm \zv_t}{\dm t} = f(t)\zv_t + \frac{g^2(t)}{2\sigma_t}\epsilonv_\theta(\zv_t,t),\quad \zv_T\sim \N(\vect{0},\tilde\sigma^2\Iv).
\end{equation}
The \emph{data prediction model} $\z_\theta(\mathbf{z}_t,t)$ predicts the original data $\z_0$ based on the noisy $\z_t$, and its relationship with $\epsilonv_\theta(\z_t,t)$ is given by $\z_\theta(\mathbf{z}_t;t) \coloneqq (\z_t - \sigma_t\epsilonv_\theta(\z_t,t)) / \alpha_t$~\citep{kingma2021variational}.
Therefore, the equivalent diffusion ODE w.r.t. the data prediction model $\zv_\theta$ is:
\begin{equation}
\label{eq:diffusion_ode_x0}
    \frac{\dm \zv_t}{\dm t} = \left(f(t) + \frac{g^2(t)}{2\sigma_t^2} \right)\zv_t - \frac{\alpha_t g^2(t)}{2\sigma^2_t}\z_\theta(\zv_t,t),\quad \zv_T\sim \N(\vect{0},\tilde\sigma^2\Iv),
\end{equation}
where the coefficients $f(t)= \frac{\dd \log \alpha_t}{\dt}$, $g^2(t)=\frac{\dd\sigma_t^2}{\dt} - 2\frac{\dd\log\alpha_t}{\dt}\sigma_t^2$~\citep{kingma2021variational}.

Given an initial value $\z_s$ at time $s>0$ and denote $\hat\z_\theta(\hat\z_\lambda,\lambda)\coloneqq \z_\theta(\z_{t_\lambda(\lambda)},t_\lambda(\lambda))$ as the change-of-variable form of $\z_\theta$ for $\lambda$, the solution $\z_t$ at time $t\in[0,s]$ of diffusion ODEs in Eq.~\eqref{eq:diffusion_ode_x0} is:
\begin{equation}
\label{eq:exact_solution_x0}
    \z_t = \frac{\sigma_t}{\sigma_s}\z_s + \sigma_t \int_{\lambda_s}^{\lambda_t} e^{\lambda} \hat\z_\theta(\hat\z_\lambda,\lambda)\dd\lambda,
\end{equation}
which can be proved by taking derivative w.r.t. $t$ in Eq.~\eqref{eq:exact_solution_x0}:
\[ \begin{aligned}
\frac{\dd \z_t}{\dd t} 
&= \frac{\dd \sigma_t}{\dd t} \frac{\z_s}{\sigma_s} 
+ \frac{\dd \sigma_t}{\dd t} \int_{\lambda_s}^{\lambda_t} e^\lambda \hat \z_\theta (\hat \z_\lambda, \lambda) \dd \lambda
+ \frac{\dd\lambda_t}{\dd t} \sigma_t e^{\lambda_t} \hat \z_\theta (\hat \z_{\lambda_t}, \lambda_t) \\
&= \frac{\dd \sigma_t}{\dd t} \frac{\z_t}{\sigma_t}
+ \frac{\dd\lambda_t}{\dd t}\sigma_t e^{\lambda_t} \hat \z_\theta (\hat \z_{\lambda_t}, \lambda_t) \\
&= \left ( f(t) + \frac{g^2(t)}{2\sigma_t^2} \right ) \frac{\z_t}{\sigma_t}
- \frac{\alpha_t g^2(t)}{2\sigma_t^2}\z_\theta (\z_{t}, t),
\end{aligned} \]
and this gives us the exact formulation as in Eq.~\eqref{eq:exact_solution_x0}. 

Based on Eq.~\eqref{eq:exact_solution_x0}, the aim of an ODE solver is to approximate the exact solution at time $t_{i}$ given the previous value $\z_{t_{i-1}}$ at time $t_{i-1}$. Denote $ \z_\theta^{(n)}(\lambda)\coloneqq \frac{\dm^n \hat \z_\theta( \z_{\lambda},\lambda)}{\dm \lambda^n}$ as the $n$-th order total derivatives of $ \z_\theta$ w.r.t. logSNR $\lambda$. ~\citet{lu2022dpm,lu2022dpm++} show that by taking the $(k-1)$-th Taylor expansion ($k\geq 1$) at $\lambda_{t_{i-1}}$ for $ \z_\theta$ w.r.t. $\lambda\in[\lambda_{t_{i-1}}, \lambda_{t_i}]$ and substitute it into Eq.~\eqref{eq:exact_solution_x0} with $s=t_{i-1}$ and $t=t_i$, we have
\begin{equation}
\label{eq:dpm_taylor_x0}
    \z_{t_i} = \frac{\sigma_{t_i}}{\sigma_{t_{i-1}}} \z_{t_{i-1}} + \sigma_{t_i}\sum_{n=0}^{k-1}
        \underbrace{%
            \vphantom{\int_{\lambda_{t_{i-1}}}^{\lambda_{t_i}} e^{\lambda}\frac{(\lambda - \lambda_{t_{i-1}})^n}{n!}\dm\lambda}
             \z^{(n)}_\theta(\hat \z_{\lambda_{t_{i-1}}}, \lambda_{t_{i-1}})}_{\mathclap{\text{estimated}}
        }
        \underbrace{\int_{\lambda_{t_{i-1}}}^{\lambda_{t_i}} e^{\lambda}\frac{(\lambda - \lambda_{t_{i-1}})^n}{n!}\dm\lambda}_{\mathclap{\text{analytically computed}}}
        + \underbrace{%
            \vphantom{\int_{\lambda_{t_{i-1}}}^{\lambda_{t_i}} e^{\lambda}\frac{(\lambda - \lambda_{t_{i-1}})^n}{n!}\dm\lambda}
            \Oc(h_i^{k+1})}_{\mathclap{\text{omitted}}},
\end{equation}
where the integral $\int e^{\lambda}\frac{(\lambda-\lambda_{t_{i-1}})^n}{n!}\dm\lambda$ can be analytically computed by integral-by-parts. Therefore, to design a $k$-th order ODE solver, we only need to estimate the $n$-th order derivatives $ \z_\theta^{(n)}(\lambda_{t_{i-1}})$ for $n\leq k-1$ after omitting the $\Oc(h_i^{k+1})$ high-order error terms. 
For $k=1$, Eq.~\eqref{eq:dpm_taylor_x0} becomes (after omitting the $\Oc(h_i^{k+1})$ terms)
\begin{equation}
\label{eq:dpm_keq1}
    \z_{t_i} = \frac{\sigma_{t_i}}{\sigma_{t_{i-1}}}\z_{t_{i-1}} + \sigma_{t_i}\z_\theta(\z_{t_{i-1}}, t_{i-1})\int_{\lambda_{t_{i-1}}}^{\lambda_{t_i}}e^{\lambda}\dd\lambda
    = \frac{\sigma_{t_i}}{\sigma_{t_{i-1}}}\z_{t_{i-1}} - \alpha_{t_i}(e^{-h_i} - 1)\z_\theta(\z_{t_{i-1}}, t_{i-1}),
\end{equation}
where $h_i\coloneqq \lambda_{t_i} - \lambda_{t_{i-1}}$ for $i=1,\dots,T$.

Since DoT is conditionally trained with partial nosing, we introduce a conditional form of Eq.~\eqref{eq:dpm_keq1} when adapting the above ODE solver into the inference stage. For $k=1$, this is given by: 
\begin{equation*}
\label{eq:condition_dpm_keq1}
    \y_{t_i} = \frac{\sigma_{t_i}}{\sigma_{t_{i-1}}}\y_{t_{i-1}} - \alpha_{t_i}(e^{-h_i} - 1)\tilde \z_\theta(\z_{t_{i-1}}, t_{i-1}),
\end{equation*}
where $\z_{t_{i-1}}=[\x;\y_{t_i-1}]$ and $\tilde \z_{\theta}(\mathbf{z}_t, t)$ is used to denote the fractions of recovered $\mathbf{z}_0$ corresponding to $\mathbf{y}_0$.

\section{Experiment Details}
\label{app:experiment-detail}
\subsection{Dataset Statistics}
We list the statistics of our used datasets in Table~\ref{tab:datasets}. For the digit multiplication datasets and GSM8K dataset, we use processed datasets from Implict CoT\footnote{\url{https://github.com/da03/implicit_chain_of_thought}}~\cite{deng2023implicit}. For boolean logic task, we construct the training and test dataset using the method from DyVal\footnote{\url{https://github.com/microsoft/promptbench/blob/main/examples/dyval.ipynb}}~\citep{zhu2023promptbench}. All datasets contain 1000 test examples except GSM8K, which contains 1319 examples.
\label{app:tab:datasets}
\begin{table}[h]
\centering
\caption{Training set size, average number of tokens in the input, intermediate, and output texts respectively when using Plaid tokenizer on the validation set and average number of rationales.}
\vskip 0.15in
\scalebox{0.9}{
\begin{tabular}{llcccc}
\toprule
\textbf{Dataset} & \textbf{Size} & \multicolumn{1}{l}{\textbf{\#Input token}} & \multicolumn{1}{l}{\textbf{\#Intermediate token}} & \multicolumn{1}{l}{\textbf{\#Output token}} & \multicolumn{1}{l}{\textbf{\#Rationales}}\\
\hline
4x4 & 808k & 16 & 84 & 15 & 4\\
5x5 & 808k & 20 & 137 & 19 & 5\\
Boolean logic & 99k & 112 & 134 & 3 & 10 \\
GSM8K-Aug & 378k & 61 & 34 & 2 &2.7 \\
\bottomrule 
\end{tabular}}
\label{tab:datasets}
\end{table}

\subsection{Details of Baselines}
When fine-tuning GPT2, we train $40$ epochs using the learning rate of \texttt{1e-4} for boolean logic and \texttt{5e-4} for others. During inference, we use greedy decoding for single decoding. For self-consistency, following the original paper~\cite{wang2022self}, we apply temperature sampling with $T = 0.5$ and truncated at the top-$k$ ($k = 40$) tokens with the highest probability for diverse generation. All GPT2-based models use \texttt{GPT2Tokenizer} with vocabulary size of $50257$. All datasets are trained using sequence length of $256$ except boolean logic, which uses $384$ length.

Note in Table~\ref{tab:main-table}, we compare Plaid DoT with the fine-tuned GPT2 small, given that the Plaid 1B~\citep{gulrajani2023likelihood} model exhibits similar perplexity to GPT2 small. This might put our Plaid DoT model at a disadvantage in terms of inference speed, as the parameters of Plaid 1B are nearly $10\times$ greater than those of GPT2 small.

For Transformer-scratch baseline~\citep{vaswani2017attention}, we use $6$ transformer encoder layers and $6$ transformer decoder layers. We employ the tokenizer from \texttt{bert-base-uncased} with a vocabulary size of $30522$. The learning rate is set to \texttt{1e-5}, and we train for 60k steps with a batch size of $128$.

For ChatGPT, we use OpenAI \texttt{api}\footnote{\url{https://platform.openai.com/docs/api-reference}} with the following prompt in $5$-shot.

\begin{table}[h]
    \centering
    \begin{minipage}{0.4\textwidth}
    \begin{tcolorbox}
    Answer the final question following the format of the given examples.\\
    
    Example problems:\\
    
    Q: \texttt{\{query\}}
    
    A: \texttt{\{answer\}}
    
    ...
    
    Question to answer:
    
    Q:

    \end{tcolorbox}
    \caption{Prompt for ChatGPT. }
    \label{tab:prompt}
    \end{minipage}
\end{table}

Please note that the throughput of ChatGPT in Table~\ref{tab:main-table} only measures the response speed of ChatGPT and does not represent the actual generation speed of the model. As a blackbox commercial product, ChatGPT may employ various optimization techniques to speedup generating responses to enhance user experiences.

\subsection{DoT Implementation Details}
\label{sec:app-imp}
We conduct all the experiments on NVIDIA V100-32G GPUs, and we use 8 GPUs for training and sampling. We resort to \texttt{half precision}~(fp16) instead of \texttt{bfloat16}~(bf16) as V100 GPU doesn't support bf16, and we don't observe any number explosion. By default, we train DoT from scratch on three datasets respectively, including the four-digit ($4\times4$), five-digit ($5\times5$) multiplication datasets, and the GSM8k dataset. Additionally, we fine-tune the pre-trained model Plaid-1B on the GSM8K dataset with DoT to explore its effectiveness further.

For DoT trained from scratch. We use $12$ layers of transformer and \texttt{bert-base-uncased} vocabulary. We preprocess the four-digit ($4\times4$) and five-digit ($5\times5$) multiplication datasets to prepare for the training process of the DoT multi-path variant, and sampling from it. The learning rate is \texttt{1e-4} and we train for 60k steps with the batch size of $128$ and max sequence length of $128$. For digit multiplication in Table~\ref{tab:main-table}, we use sampling step $T=1$ to achieve high throughput while keeping the accuracy. For boolean logic dataset, we use $T=2$.

For DoT fine-tuned from Plaid, we set the training steps of the DoT and multi-pass DoT to be 120k and 30k respectively, as we find more training steps will lead to performance degradation. The learning rate is set to \texttt{1e-4} for boolean logic and \texttt{3e-4} for other datasets. The max sequence length is set to $384$ for the boolean logic dataset and $256$ for others. We use Adam optimizer~\citep{Kingma2014AdamAM}. During tokenization, we use Plaid's tokenizer and we treat all the digits as individual tokens. During training, we set $\epsilon_{min}$ to be 0.95 as we find decreasing the probability of oracle demonstration hinders model training. We choose glancing sampling $\gamma=0.01$ and self consistency $m=20$. Following~\citet{gulrajani2023likelihood}, we also adopt self-conditioning~\citep{chen2022analog} during training.  During inference, we set the scoring temperature to 0.5 to sharpen the predicted noise distribution. We also use soft logits with a temperature of $0.5$ to produce more diverse samples. By default, we use sampling step $T=64$ to ensure accuracy. Training DoT and \ourmodeldot require 29h and 10h, respectively.
For DoT trained from SEDD, we set the training steps of the DoT and multi-pass DoT to be 200k, with other parameters being the same as when training Plaid. 
For all the experiments, we have verified the statistical significance by running them multiple times.

\subsection{Additional Results}
\paragraph{Comparison to larger open language models.}
\begin{table}[ht]
\centering
\caption{Comparison to larger AR models.}
\begin{tabular}{llr}
\toprule
\textbf{} & \textbf{Params} & \textbf{Accuracy} \\
\hline
GPT-2-medium CoT & 355M & 43.9 \\
Mistral CoT & 7B & 68.8 \\
Llama CoT & 7B & 59.0 \\
SEDD-medium \ourmodelmpdot & 424M & 53.5 \\
\bottomrule
\end{tabular}
\label{app-tab:llms}
\end{table}
We compare our model with LoRA fine-tuning of AR LLMs on the same GSM-Aug dataset, which is listed in Table~\ref{app-tab:llms}. Please note that the current diffusion pretrained model is much smaller than Llama 7B, so this comparison is not fair and we just list them for reference. We have validated that our DoT is better than the same scale autoregressive model GPT-2, which shares a similar architecture with Llama. We believe that further exploration of diffusion language models will lead to larger models that can compete with current LLMs, allowing DoT to achieve results more comparable to Llama.

\paragraph{Comparison with no-DoT finetune.}
\begin{table}[ht]
\centering
\caption{Comparison between DoT and no-DoT (Answer-only).}
\begin{tabular}{lr}
\toprule
\textbf{} & \textbf{Accuracy} \\
\hline
GPT-2-small Answer-only & 13.3 \\
GPT-2-small CoT & 39.0 \\
Plaid Answer-only & 12.4 \\
Plaid \ourmodelmpdot  & 37.7 \\
SEDD-small Answer-only & 29.1 \\
SEDD-small \ourmodelmpdot  & 43.2 \\
\bottomrule
\end{tabular}
\label{app-tab:nodot}
\end{table}
We conduct the answer-only setting to further validate the effectiveness of DoT. The results in Table~\ref{app-tab:nodot} reveal that fine-tuning diffusion models solely with answer data leads to inferior performance compared to DoT, mirroring the degradation of AR models in the absence of CoT.

\paragraph{Throughput Comparison.}
\begin{table}[ht]
\centering
\caption{Throughput comparison when increasing the number of timesteps $T$ for Plaid \ourmodelmpdot.}
\begin{tabular}{lrr}
\toprule
\textbf{T} & \textbf{Accuracy} & \textbf{Throughput} \\
\hline
1 & 18.2 & 6.6 \\
2 & 35.9 & 3.4 \\
4 & 36.7 & 1.7 \\
8 & 36.4 & 0.9 \\
16 & 36.1 & 0.4 \\
32 & 37.4 & 0.2 \\
64 & 37.7 & 0.1 \\
128 & 37.7 & .05 \\
\bottomrule
\end{tabular}
\label{app-tab:throughput}
\end{table}
We have shown how T affects performance on grade school math in Figure \ref{fig:timestep-t}, and here we also show how T affects throughput for Plaid \ourmodelmpdot, as in Table~\ref{app-tab:throughput}. The relationship between throughput and T appears to be nearly linear.

\paragraph{Comparison of the reasoning paths between DoT and \ourmodelmpdot.}
We observe that \ourmodelmpdot outperforms DoT in correctness regarding the reasoning paths, while DoT slightly excels in diversity as depicted in Figure \ref{fig:reasoning-efficiency}(b). Below we show some examples where \ourmodelmpdot can predict the correct reasoning path while DoT fails:
\begin{tcolorbox}[breakable]
\textbf{Query}: The Kennel house keeps 3 German Shepherds and 2 Bulldogs. If a German Shepherd consumes 5 kilograms of dog food and a bulldog consumes 3 kilograms of dog food per day. How many kilograms of dog food will they need in a week?

\textbf{DoT}: <<3$*$5=15>> <<7$*$3=21>> <<15+21=36>> \#\#\#\# 36

\textbf{\ourmodelmpdot}
: <<3$*$5=15>> <<2$*$3=6>> <<15+6=21>> <<21$*$7=147>> \#\#\#\# 147
\end{tcolorbox}

\begin{tcolorbox}[breakable]
\textbf{Query}: Skyler has 100 hats on his hand with the colors red, blue, and white. Half of the hats are red, 3/5 of the remaining hats are blue, and the rest are white. How many white hats does Skyler have?

\textbf{DoT}: <<1/2$*$100=50>> <<3/5$*$50=30>> <<100-30=70>> \#\#\#\# 70

\textbf{\ourmodelmpdot}
: <<100/2=50>> <<100-50=50>> <<50$*$3/5=30>> <<50-30=20>> \#\#\#\# 20
\end{tcolorbox}

\subsection{Other Attempts}
For the ablation design for DoT fine-tuning in Table~\ref{tab:ablation-gsm}, we have tried to fine-tune a decoder-only autoregressive language model (i.e., GPT2 here), where we only change the base model from Plaid 1B to GPT2 large, remove the causal mask and keep all other diffusion training settings the same with the Plaid fine-tuning. In this setting, even though the model is formulated and trained in the diffusion manner, it still can not predict the right format of answers. This experiment may indicate that a pre-trained diffusion model is necessary for the further fine-tuning of downstream tasks.

Regarding datasets, we also try to mix up four-digit ($4\times4$) and five-digit ($5\times5$) multiplication datasets for training and testing, considering that the number of rationales is different in these two tasks. As for the result, the trained model learns when to conclude the computation and can attain 100\% accuracy.

\section{Discussion about base models}
\label{sec:app-main-tab}
Our DoT approach is constrained by the pre-training and fine-tuning paradigm due to the not-strong-enough base models. This lags behind the current trend of instruction-tuning LLMs and pursuing the generalization of LMs across various tasks. Nevertheless, considering the pre-trained diffusion models are still in their early stages and the lack of scaled pre-trained diffusion models, our study is a preliminary exploration to show the potential of diffusion models for reasoning tasks, and we believe that with more powerful pre-trained diffusion models and post-instruction tuning, DoT can attain the generalization capabilities of today's LLMs and yield further advantages. 

\section{Boarder Impacts}
\label{app:boarder-impacts}
Our work contributes to the understanding of denoising generative models and enhances their generation capabilities within certain discrete text reasoning datasets. 
The proposed DoT with diffusion language models challenges autoregressive models with CoT, achieving competitive performance. While there is still a large gap with modern large autoregressive language models such as ChatGPT, we believe DoT can benefit more with future work on scaling diffusion language models.
However, we acknowledge that deep generative models, as powerful tools for learning from unstructured data, can have detrimental societal impacts if misused. Specifically, these models can facilitate the spread of misinformation by reducing the resources required to create realistic fake content. Additionally, the generated samples from these models accurately reflect the statistics of their training datasets. Consequently, if these samples are interpreted as objective truth without considering the inherent biases present in the original data, they can perpetuate discrimination against minority groups.

\end{document}